\documentclass[journal]{IEEEtai}

\usepackage{color,array}
\usepackage{graphicx}

\usepackage{balance}
\usepackage{ragged2e}
\usepackage{color}
\usepackage{tikz}
\usepackage{textcomp}
\usepackage[numbers]{natbib}
\usepackage{amsfonts}
\usepackage{comment}
\usepackage{caption}
\usepackage{subcaption}
\usepackage{array}
\usepackage{tabularx}
\usepackage{multirow}
\usepackage{mathtools}
\usepackage{algorithm2e}
\usepackage{amssymb}

\begin{document}

\title{{Facilitating Sim-to-real by Intrinsic Stochasticity of Real-Time Simulation in Reinforcement Learning} for Robot Manipulation}

\author{Amir M. Soufi Enayati$^{1\dagger}$,~\IEEEmembership{Student~Member~IEEE}, Ram Dershan$^{2\dagger}$, Zengjie Zhang$^{2}$,~\IEEEmembership{Member~IEEE},\\ Dean Richert$^{2}$,~\IEEEmembership{Member~IEEE}, and Homayoun Najjaran$^{1*}$,~\IEEEmembership{Member~IEEE}

\thanks{This work receives financial support from Kinova\textregistered~Inc. and Natural Sciences and Engineering Research Council (NSERC) Canada under the Grant CRDPJ 543881-19. We also benefit from the hardware resources for the experimental setup provided by Kinova\textregistered~Inc.}

\thanks{$^1$Amir M. Soufi Enayati and Homayoun Najjaran are with Faculty of Engineering and Computer Science, University of Victoria, 3800 Finnerty Road, Victoria, BC, V8P 5C2, Canada, {\tt\small \{amsoufi, najjaran\}@uvic.ca}.}

\thanks{$^2$Ram Dershan, Zengjie Zhang, and Dean Richert are with School of Engineering, The University of British Columbia, 1137 Alumni Avenue, Kelowna, BC, V1V 1V7, Canada, {\tt\small \{ram.dershan, zengjie.zhang, dean.richert\}@ubc.ca}.}

\thanks{$\dagger$~Equivalent contribution as the common first authors.}

\thanks{*~Corresponding author.}
}


\maketitle

\begin{abstract}
\justifying
Simulation is essential to reinforcement learning (RL) before implementation in the real world, especially for safety-critical applications like robot manipulation. Conventionally, RL agents are sensitive to the discrepancies between the simulation and the real world, known as the sim-to-real gap. The application of domain randomization, a technique used to fill this gap, is limited to the imposition of heuristic-randomized models. {We investigate the properties of intrinsic stochasticity of real-time simulation (RT-IS) of off-the-shelf simulation software and its potential to improve RL performance. This improvement includes a higher tolerance to noise and model imprecision and superiority to conventional domain randomization in terms of ease of use and automation. Firstly, we conduct analytical studies to measure the correlation of RT-IS with the utilization of computer hardware and validate its comparability with the natural stochasticity of a physical robot. Then, we exploit the RT-IS feature in the training of an RL agent. The simulation and physical experiment results verify the feasibility and applicability of RT-IS to robust agent training for robot manipulation tasks. The RT-IS-powered RL agent outperforms conventional agents on robots with modeling uncertainties. RT-IS requires less heuristic randomization, is not task-dependent, and achieves better generalizability than the conventional domain-randomization-powered agents. Our findings provide a new perspective on the sim-to-real problem in practical applications like robot manipulation tasks.}
\end{abstract}

\begin{IEEEImpStatement}
Roboticists face a challenge with the sim-to-real problem as direct learning in physical systems is unsafe and costly, while simulations often lack the fidelity required for deployment. This paper presents an automated solution, in contrast with the heuristics-based conventional domain randomization methods. We introduced real-time intrinsic stochasticity (RT-IS) in a physics-based simulation engine to randomize the model. First, the adequacy of this approach is validated by quantifying the resulting deviation and comparing it with our physical setup. Second, our experiments indicated that RL agents trained with RT-IS outperform other agents when transferred to the real robot. The results demonstrate the potential to minimize the need for expert design in manipulation tasks and decrease the sensitivity of RL agents to simulation discrepancies. We conclude that emulating noise in a physically plausible manner is critical to mitigating model mismatch concerns and implementing RL agents in real robotic systems.
\end{IEEEImpStatement}


\section{Introduction}
\IEEEPARstart{T}{oday}, the world finds itself amid an exciting industrial revolution represented by \emph{Industry 4.0}, a paradigm that has pushed manufacturing systems to become more intelligent, reliable, and efficient by utilizing innovative technologies including artificial intelligence (AI), cyber-physical systems, internet of things, and cloud computing~\cite{zhong2017intelligent, zhang2022}. AI technologies provide the tools needed to build intelligent manufacturing systems that are flexible, smart, and easily re-configurable. AI has applications in just about any manufacturing process such as process monitoring~\cite{dalzochio2020machine}, process optimization~\cite{morariu2020machine}, and process control~\cite{chin2020machine}. AI enables manufacturing devices and machines, especially robots, to self-monitor and autonomously respond to different situations and environmental changes~\cite{pires2021role}. The next generation of intelligent manufacturing requires the robots to be capable of changing their behaviors and modals automatically adapted to the peripheral changes without human intervention or manual reprogramming~\cite{SOUFIENAYATI2022381}. 

Reinforcement Learning (RL) is one of the most promising technologies that can achieve this goal for robotic systems. The RL framework introduces autonomous decision-making based on a feedback and reward setting, where an agent applies actions to an environment according to certain observations. The applied actions are solved subject to the optimization of predefined rewards. Different from the conventional control methods built on heuristics, RL does not usually require prior knowledge of the environment, which is also referred to as \textit{model-free}~\cite{tutsoy2021model}. Instead, the RL training process takes a trial-and-error approach to learn the best commands for a specific task, such as grasping \cite{zhang2021robot}, navigation \cite{zielinski20213d}, and Braille typing \cite{church2020deep}. This framework allows a robot to learn a feasible control solution automatically from the interaction data without manually programming the robot and task models. To provide sufficient data for the training of RL, simulation of the system is used to save the temporal and hardware resources. Nevertheless, due to the mismatch of the features between the physical systems and their simulation models, the agents trained in simulation do not always reproduce their performance in the real world, which is also referred to as the sim-to-real gap~\cite{kober2013reinforcement} or the \emph{reality gap}~\cite{8202133}. The conventional RL methods are sensitive to the variation between the virtual and the physical robot dynamics~\cite{8202133} caused by the reality gap, which becomes a major barrier preventing the application of RL to practical problems and remains an open question that prevents the wide application of RL to practice.

An inspiring idea to fill the sim-to-real gap is found in robust control, a well-known concept in the conventional control theory to cope with the existence of system uncertainties~\cite{petersen2014robust}. Robust control aims at designing a control method that is feasible for all system models that are defined in a closed model set that covers all possible features of the real system. {Thus, a similar concept namely \textit{Robust RL} is proposed as a different avenue taken to tackle the issue of uncertainty and disturbance in real applications of RL. The core idea was an amalgamation of the robust control law of $H_\infty$ and reinforcement learning, suggested by~\cite{morimoto2005}. In this approach, the disturbance is modeled as an agent with a non-deterministic but bounded range of \textit{disturbance} trying to destabilize the RL agent. The disturbance model impedes the controller by applying the biggest model discrepancy or input noise. In robust \textit{adversarial} RL, the agent is equipped with another deep learning model which learns the extent of its disturbance output throughout the training process~\cite{pinto2017}. Various domains of action for the mentioned adversary have also been experimented on, spanning robustness to transition (environment dynamics), disturbance (external unmodeled forces), action (input noise), and observation (transducer uncertainty)~\cite{moos2022}.
{The application of Robust RL formulation has been expanded, in a very diverse set of methods, much further than sim-to-real transfer in recent years. In another work, a robust goal-conditioned reinforcement learning approach for end-to-end robotic control in adversarial and sparse reward environments is introduced~\cite{he2023}. Their system utilizes a mixed adversarial attack scheme to generate diverse perturbations on observations and employs a hindsight experience replay technique to transform failed experiences into successful ones. Also, proposed by~\cite{he2022robust}, an observation-based adversarial reinforcement learning approach for robust lane change of autonomous vehicles is presented, addressing the challenges of perception uncertainty and adversarial observation perturbations. At a more fundamental level, \cite{panaganti2022robust} devises an efficient way to exploit perturbed historical data from adversarial environments. This can especially be beneficial in the case of sim-to-real where generating perturbed simulated environments is affordable. Furthermore, a model-based reinforcement learning algorithm is devised by~\cite{panaganti2022sample} for learning an $\epsilon$-optimal robust policy where the nominal model is unknown. This algorithm considers different forms of uncertainty sets and provides precise characterizations of sample complexity by generative models. This provides an avenue to a range of novel robust RL algorithms.} The intuition behind the robust RL concept is astute and it suggests certain guarantees of stability, however, structuring such problems requires two sources of prior knowledge: choosing the domains of impact for the adversary and designing the extent of said impact. In many cases, this information is not provided before the actual implementation. Mismodeling in this design may lead to over- or under-compensation of the discrepancies and sub-optimal performance. Additionally, parameter tuning the agents equipped with this level of sophistication can be disinclined in the industrial use of the technology.} 

\begin{figure}[htbp]
\centering
\noindent

\begin{tikzpicture}[scale=1,font=\small]

\def\nw{2cm}
\def\nh{1cm}
\def\cn{0.15cm}

\definecolor{s_pink}{RGB}{255, 153, 153}
\definecolor{s_blue}{RGB}{153, 204, 255}
\definecolor{s_yellow}{RGB}{255, 230, 153}

\node[minimum height=\nh,minimum width=\nw, text width=\nw,align=center,draw, thick,fill=s_yellow, rounded corners=\cn] (sim_env) at (0cm,0cm) {{\footnotesize \textbf{\color{red} (1)}} Simulation Environment};
\node[minimum height=\nh,minimum width=\nw, text width=\nw,align=center,draw, thick,fill=s_blue, rounded corners=\cn] (sim_pi) at (0cm,1.5cm) {Trained Policy};

\draw[->,>=stealth,thick] (sim_pi.east) -- ([xshift=0.3cm] sim_pi.east) -- node[pos=0.5,align=right, anchor=east]{action} ([xshift=0.3cm] sim_env.east) -- (sim_env.east);
\draw[->,>=stealth,thick] (sim_env.west) -- ([xshift=-0.3cm] sim_env.west) -- node[pos=0.5,align=left, anchor=west]{state} ([xshift=-0.3cm] sim_pi.west) -- (sim_pi.west);

\draw[->,>=stealth,very thick,dashed, color=red] ([yshift=-0.5cm] sim_env.south) -- node[pos=0.3,align=left, anchor=west]{\footnotesize \textbf{\color{red} (3)}}(sim_env.south);

\node[minimum height=\nh,minimum width=\nw, text width=\nw,align=center,draw, thick,fill=s_pink, rounded corners=\cn] (real_env) at (4cm,0cm) {Real Environment};
\node[minimum height=\nh,minimum width=\nw, text width=\nw,align=center,draw, thick,fill=s_blue, rounded corners=\cn] (real_pi) at (4cm,1.5cm) {Deployed Policy};

\draw[->,>=stealth,thick] (real_pi.east) -- ([xshift=0.3cm] real_pi.east) -- node[pos=0.5,align=right, anchor=east]{action} ([xshift=0.3cm] real_env.east) -- (real_env.east);
\draw[->,>=stealth,thick] (real_env.west) -- ([xshift=-0.3cm] real_env.west) -- node[pos=0.5,align=left, anchor=west]{state} ([xshift=-0.3cm] real_pi.west) -- (real_pi.west);

\draw[->,>=stealth,very thick,dashed, color=red] ([yshift=0.5cm] real_pi.north) -- node[pos=0.3,align=left, anchor=west]{\footnotesize \textbf{\color{red} (2)}}(real_pi.north);

\draw[<->,>=stealth,dashed, double, thick] ([xshift=0.5cm, yshift=-0.8*\nh] sim_pi.east) -- node[pos=0.5,align=center, anchor=south, text width=1cm]{Reality Gap} ([xshift=-0.5cm, yshift=-0.8*\nh] real_pi.west) ;

\end{tikzpicture}
\caption{ The illustration of the \textit{reality gap} and the three common approaches to solving it: (1) system identification, (2) domain adaptation, and (3) domain randomization.}
\label{fig:framework}
\end{figure}
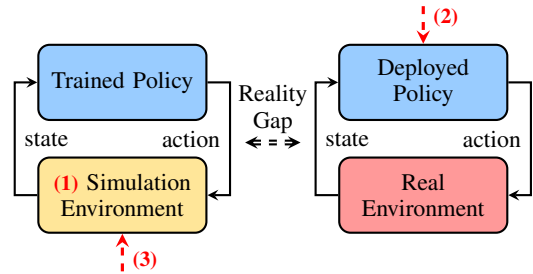

The existing solutions for RL agents to alleviate the sim-to-real problem mainly include system identification~\cite{aastrom1971system}, domain adaptation~\cite{bousmalis2018domap}, domain randomization~\cite{tobin2017domain}. {The relation between these three solutions and the reality gap is illustrated in Fig.~\ref{fig:framework}.}
\textit{System identification} aims to solve the sim-to-real gap by mathematically estimating a specific pre-structured model based on data~\cite{jv2021sysid}, e.g., by defining a regression problem~\cite{kaess2008isam, ding2016recursive}. This approach is most effective for tuning time-invariant parameters, however in reality a major problem is noise which is of a different nature. Some recent cases of implementation of machine learning for modeling time-variant parameters, however, can be found~\cite{niu2022deep, khodabandehlou2018nonlinear}. Despite these results, the efficiency of system identification, in this case, is questionable due to the requirement of tremendous data and exhaustive modeling efforts. \textit{Domain adaptation} is inspired by the computer vision field, where the data from a \emph{source domain} is used to improve the performance on a different \emph{target domain}~\cite{bousmalis2018domap}. It has been used as the critical technology of transfer learning~\cite{pan2009survey}. From the perspective of sim-to-real, the source is the simulation environment and the real world serves as the target. This technique focuses on the transformation of feature space either by using statistical-discrepancy methods~\cite{long2015learning} or adversarial-learning methods~\cite{jiang2021simgan} such as generative adversarial networks (GAN). Then, the model for the target domain can be generated by adapting the source model utilizing the feature transformation. However, the application of domain adaptation to sim-to-real is hindered by the fact that it requires data from both domains.

Another solution to mitigate the sensitivity of RL is domain randomization, also known as \textit{dynamic variability}~\cite{12523627020170815}, which promotes the generalizability of an RL agent by exciting with multiple varying model configurations~\cite{8460528, wang2020reinforcement}. The main technical point of domain randomization is to impose heuristic noise to the models or observations of the environment during the training process~\cite{plappert2017parameter, igl2019generalization}, such that the agents can achieve a balanced performance over various models and observations. The spaces of the perturbed models and the observations are referred to as \textit{domain}s, and the manner of imposing heuristic noise in these domains is referred to as \textit{randomization}. The randomization can be of any two types visual~\cite{tobin2017domain} and dynamic~\cite{peng2018sim}. Visual randomization is mostly related to computer vision tasks such as object detection~\cite{eversberg2021generating, moreu2022domain}, object recognition~\cite{Du2021recog}, or pose estimation~\cite{hagelskjaer2022parapose}. Dynamic randomization targets robotics and control tasks by randomizing dynamic features such as mass, friction, and measurement noise in observations. The technology of domain randomization can avoid the overfitting of an RL agent to a single environment model and guarantee its generalizability to all perturbed environments~\cite{igl2019generalization}. Similar technology is also used to avoid local optimums~\cite{lillicrap2015continuous, plappert2017parameter}. However, there exists a trade-off between the extent of randomization and the performance of the agent. Firstly, the randomization sacrifices the training performance of the agent on individual environment models. Thus, large-extent of randomization may not pay off the overall loss of the training performance. Secondly, randomization increases the variance of the trained policy, which makes the test performance unstable and difficult to predict. Thirdly, similar to robust control, large-extent domain randomization may lead to the infeasibility of the agent training, i.e., the failure of policy convergence. Therefore, determining the extent of the heuristic randomization is always a challenge.

{Naturally, the approaches to tackle the challenge targeted by this paper are not limited to the above. For instance, a hierarchical combination of RL (high-level) and  precise conventional controller (low-level) is proposed by~\cite{shahid2022continuous}. The RL agent creates setpoints at a lower rate for a stiff controller in the simulation. The conventional controller can compensate for the disturbances, isolating the RL agent from uncertainties. It has also been shown that decentralizing the controllers minimizes the effect of uncertainty~\cite{shahid2021decentralized}. It should be noted that delays can form another source of challenge for sim-to-real, besides noise and disturbance~\cite{dulac2021challenges}. These delays can be caused because of the low update rate of the RL agent on real hardware, the latency of the action execution in the actuators, or both. In~\cite{ramstedt2019real}, adding an estimator to predict the state changes during action selection improved the performance in real-time. Even the mere presence of random delays has shown a general improvement in the robustness of the RL agent~\cite{bouteiller2021reinforcement, ji2022hierarchical}. The latter inspires the authors to verify the hypothesis of the benefit of a source of intrinsic noise to robustness from a different perspective. Unlike delays that follow the dynamics of the system or the actuator, action/observation noise is less predictable or possibly not at all. Therefore, we propose using a meaningful source of noise in simulation that can be related to the same phenomenon in the physical system.}

In the preliminaries of this work, we find that the intrinsic stochasticity of real-time simulation (RT-IS) is promising to provide less-heuristic solutions to domain randomization, although it has not been widely discussed in the previous work. Real-time simulation is a special mode of off-the-shelf simulation software, such as like PyBullet. The simulation step time in the real-time mode is stochastically changed due to the uncertain resource management of the operating system, which leads to the stochasticity of the entity states of the simulation, such as the movement trajectories of the robot. Such stochasticity is \textit{intrinsic} since it is affected by the internal computing resources, such as $CPU$, $GPU$, and $Memory$ usage, similar to the natural stochasticity of physical robotic systems due to the change of temperature, humidity, and lubricant conditions~\cite{asada2011special}. 

This naturally leads us to the idea to exploit RT-IS to improve the robustness of a conventional RL agent and reduce the heuristics of a conventional domain-randomized agent. To evaluate the feasibility of this solution, the following {three} questions should be answered.
\begin{enumerate}
\item Does the utility of hardware resources affect RT-IS?
\item Is RT-IS comparable to the physical robot?
\item Can RT-IS be used to improve the robustness of RL and fill the sim-to-real gap?
\end{enumerate}
The answers can help rediscover the functionality of RT-IS which is promising to promote the fidelity of robot simulation~\cite{keesman2011system} and improve the robustness of RL methods~\cite{polydoros_nalpantidis_2017}.

{In this paper, we give attempt to fill the sim-to-real gap by answering the three questions above related to RT-IS}, which has never been investigated by the existing research work, according to our knowledge. {Specifically, through statistical studies, we investigate the cause of RT-IS in an off-the-shelf simulation environment and validate its similarity to the stochastic uncertainty of the physical robot. This provides the practical foundation for utilizing RT-IS to replicate the randomness of real robots and fill the sim-to-real gap. Then, we propose a method that utilizes domain randomization to render robust RL agents. Comparison studies of an essential point-to-point robot manipulation task are conducted to validate the equivalence between RT-IS and domain randomization for RL agent training.} 

The results of this paper are promising to provide a more realistic and implementable solution for the practical applications of RL methods to robot manipulation tasks with the existence of sim-to-real gaps, {i.e., utilizing RT-IS to train a robust RL agent instead of domain randomization, with fewer computation resources}. The remaining part of the paper is organized as follows. Sec.~\ref{sec:data_collect} introduces our experimental setup in simulation and on the physical robot for the data collection for the study of RT-IS. In Sec.~\ref{sec:result}, we analyze the influence of CPU, GPU, and Memory consumption on the intrinsic stochasticity and evaluate the magnitude of stochasticity between the real-time simulation and the real robot. Sec~\ref{sec:exp} explains the configuration of our experimental studies to evaluate the effectiveness of RT-IS. The experimental results are evaluated and analyzed in Sec.~\ref{sec:exp_result}. Finally, Sec.~\ref{sec:con} concludes the paper.

\section{Data Collection For Analytical Study}
\label{sec:data_collect}

{This section introduces the collection process of the data used for the correlation analysis. We first interpret the experimental setup and the data collection in the real-time simulation environment. Then, we explain the data collection process on the physical robot. The overall process is illustrated in Fig.~\ref{fig:data_collect}. The identical heuristic policy based on a constant-speed controller is applied to both the simulation and the physical robot. The simulation is performed in the real-time mode where the output $\tilde{V}$ is affected by RT-IS. The output of the physical robot $V$ is affected by the physical stochastic uncertainties.}

{
\begin{figure}[htbp]
\centering
\noindent

\begin{tikzpicture}[scale=1,font=\small]

\def\nw{2cm}
\def\nh{1cm}
\def\cn{0.15cm}

\definecolor{s_pink}{RGB}{255, 153, 153}
\definecolor{s_blue}{RGB}{153, 204, 255}
\definecolor{s_yellow}{RGB}{255, 230, 153}

\node[minimum height=\nh,minimum width=\nw, text width=\nw,align=center,draw, thick,fill=s_yellow, rounded corners=\cn] (sim_env) at (0cm,0cm) {Real-Time Simulation};
\node[minimum height=\nh,minimum width=\nw, text width=\nw,align=center,draw, thick,fill=s_blue, rounded corners=\cn] (sim_pi) at (0cm,1.5cm) {Heuristic Policy};

\draw[->,>=stealth,thick] (sim_pi.east) -- ([xshift=0.3cm] sim_pi.east) -- node[pos=0.5,align=right, anchor=east]{$\tilde{T}$} ([xshift=0.3cm] sim_env.east) -- (sim_env.east);
\draw[->,>=stealth,thick] (sim_env.west) -- ([xshift=-0.3cm] sim_env.west) -- node[pos=0.5,align=left, anchor=west]{$\tilde{V}$} ([xshift=-0.3cm] sim_pi.west) -- (sim_pi.west);

\draw[->,>=stealth,very thick,dashed] ([yshift=-0.5cm] sim_env.south) -- node[pos=0,align=left, anchor=north]{\footnotesize \textbf{RT-IS}}(sim_env.south);

\node[minimum height=\nh,minimum width=\nw, text width=\nw,align=center,draw, thick,fill=s_pink, rounded corners=\cn] (real_env) at (4cm,0cm) {Physical Robot};
\node[minimum height=\nh,minimum width=\nw, text width=\nw,align=center,draw, thick,fill=s_blue, rounded corners=\cn] (real_pi) at (4cm,1.5cm) {Heuristic Policy};

\draw[->,>=stealth,thick] (real_pi.east) -- ([xshift=0.3cm] real_pi.east) -- node[pos=0.5,align=right, anchor=east]{$T$} ([xshift=0.3cm] real_env.east) -- (real_env.east);
\draw[->,>=stealth,thick] (real_env.west) -- ([xshift=-0.3cm] real_env.west) -- node[pos=0.5,align=left, anchor=west]{$V$} ([xshift=-0.3cm] real_pi.west) -- (real_pi.west);

\draw[->,>=stealth,very thick,dashed] ([yshift=-0.5cm] real_env.south) -- node[pos=0,align=left, anchor=north]{\footnotesize \textbf{Physical Uncertainties}}(real_env.south);

\node[minimum height=0.6*\nh,minimum width=1.5*\nw, text width=1.5*\nw,align=center,draw, thick,fill=white, rounded corners=\cn] (data_c) at (2.6cm,2.5cm) {Data Collection};

\draw[->,>=stealth] ([yshift=-0.2cm] sim_env.west) -- ([xshift=-0.5cm, yshift=-0.2cm] sim_env.west) -- ([xshift=-0.5cm, yshift=1cm] sim_pi.west) -- node[pos=0.46,align=left, anchor=south]{\footnotesize CPU/GPU/Memory}(data_c.west);

\end{tikzpicture}
\caption{ The illustration of the experimental setup and the data to be collected, where $V$ and $T$ are the joint velocity and torque of the physical robot, and $\tilde{V}$ and $\tilde{T}$ are those of the virtual robot in the simulation environment.}
\label{fig:data_collect}
\end{figure}
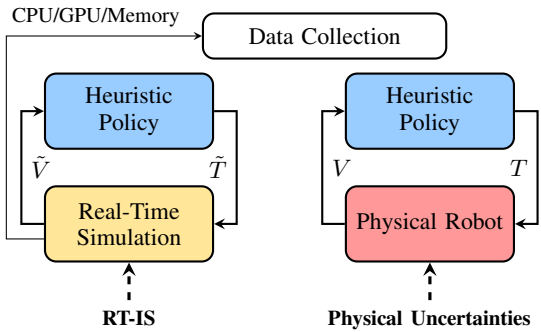

}

\subsection{Experimental Setup in Real-Time Simulation}
\label{sec:sim}

We choose PyBullet, an open-source simulator based on the Bullet physics engine for Python, as our simulation environment. PyBullet offers parallel simulation functionality, compatibility with Gym and other machine learning libraries, and high-speed low-demand calculations that are favorable for reinforcement learning research. The stability and viability of PyBullet in different robotic applications (e.g., rigid body motion, locomotion, mobile robots, collision, and contact) are widely recognized \cite{korber2021comparing}. PyBullet supports the real-time simulation mode based on the real-time calling mechanism of the operating system, which provides superior fidelity to reality but brings up stochastic uncertainty of simulation. Nevertheless, the real-time simulation mode of PyBullet is the main interest of this paper. To run the simulation, we use a powerful workstation that is capable of training reinforcement learning agents with high computational capabilities. The workstation is equipped with an AMD Ryzen 9 5900X CPU, an Nvidia RTX 3090 GPU, and 64GB (2x32) Corsair DDR4 memory, which is superior to many papers on simulation benchmarks \cite{ayala2020comparison, pitonakova2018feature}. Moreover, the operating system is Ubuntu 18.04.6 LTS. The purpose of this setup is to validate whether ordinary simulation tools on powerful computers are affected by minor changes in resource consumption.

In the PyBullet simulation, we render a virtual model of a six-degree-of-freedom (6DoF) Kinova\textregistered~Gen3 robot with a Robotiq\textregistered~2F-85 gripper, in a Gym environment. The robot model is developed from the official URDF file provided by the manufacturer. We enabled the real-time simulation timesteps option using PyBullet's \emph{setRealTimeSimulation()}. This feature, in contrast with discrete-time increments, does not deal with time as a deterministic parameter. Instead, the model is updated according to the machine's internal clock as the reference for timesteps. While deterministic timesteps are regularly used to take out unpredictability, this feature involves the stochastic nature of digital computation in the simulation. The simulated robot is manipulated in position control mode with a standard constant-coefficient Proportional-Integrator-Derivative (PID) controller. Implementing a common low-order controller guarantees a limited calculation. To restrict other sources of unpredictable behavior, all other kinematic and dynamic equations are solved in the forward direction (from joint space parameters to task space) and hence, needless of any random seed initialization.

\subsection{Data Collection in Real-Time Simulation}
\label{sec:datasim}

\begin{figure}[htbp]
\centering
     \begin{subfigure}[b]{0.48\textwidth}
         \centering
         \includegraphics[width=0.96\textwidth, trim={0 2cm 0 0},clip]{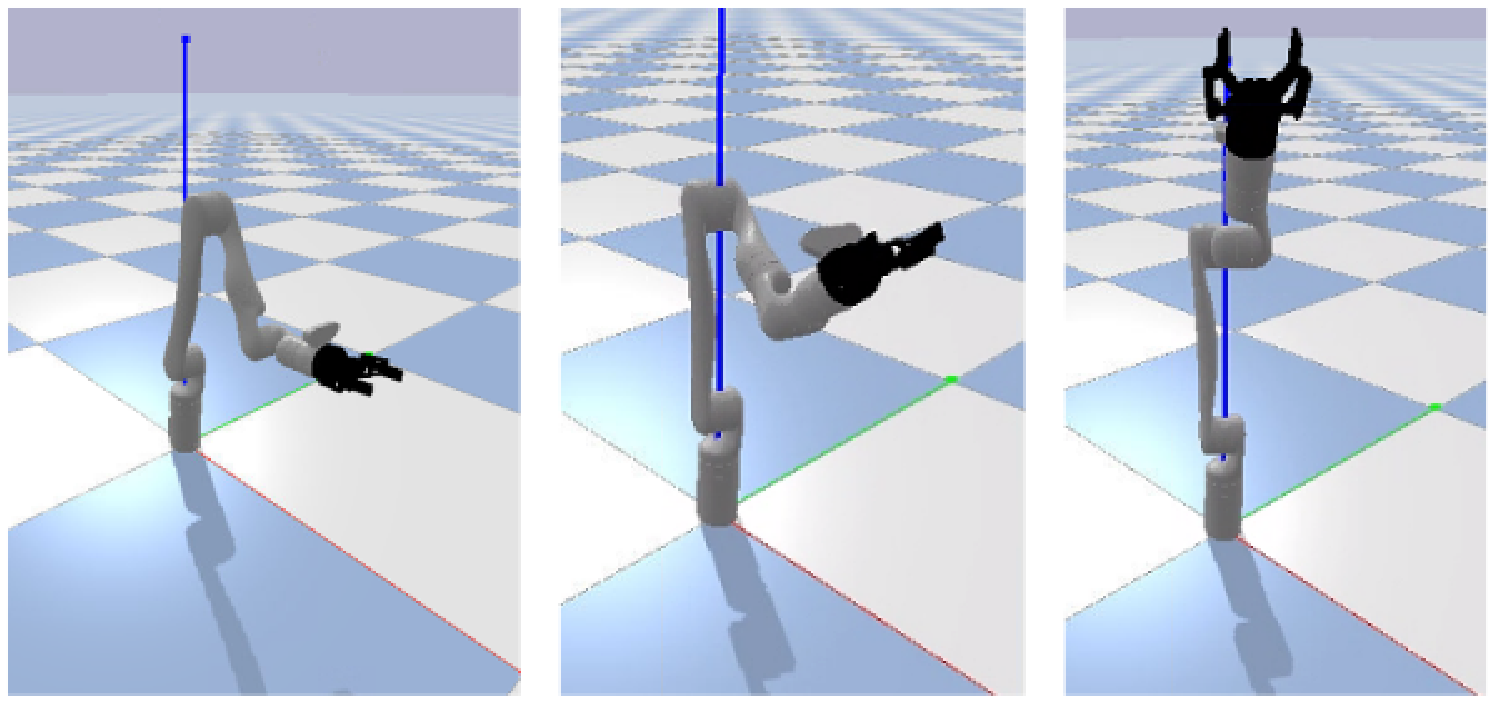}
         \caption{The P2P reaching task}
         \label{fig:p2p}
     \end{subfigure}\vspace{1mm}
     \begin{subfigure}[b]{0.48\textwidth}
         \centering
         \includegraphics[width=0.96\textwidth, trim={0 1.5cm 0 1.5cm},clip]{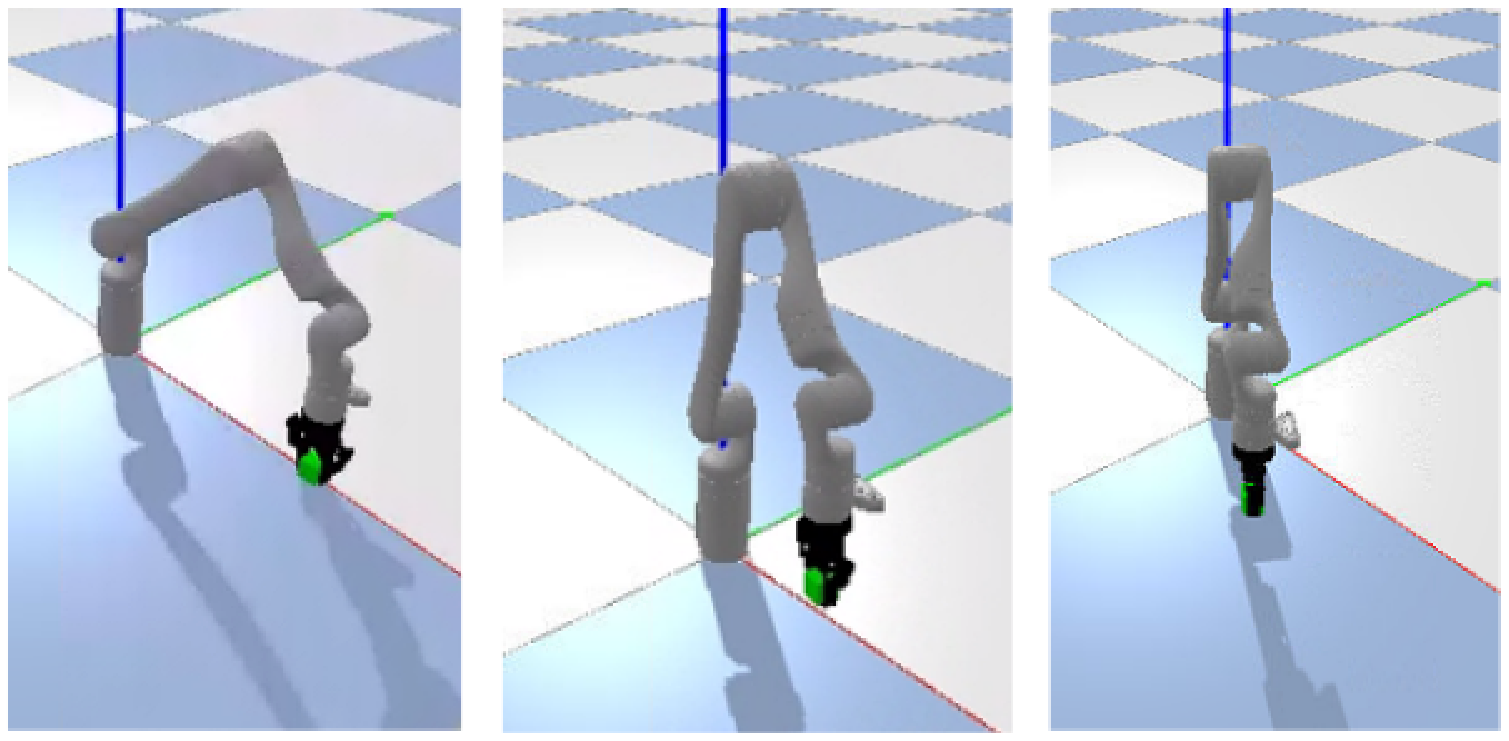}
         \caption{The P\&P task}
         \label{fig:pnp}
     \end{subfigure}\vspace{1mm}
     \begin{subfigure}[b]{0.48\textwidth}
         \centering
         \includegraphics[width=0.95\textwidth, trim={0 1.7cm 0 1.3cm},clip]{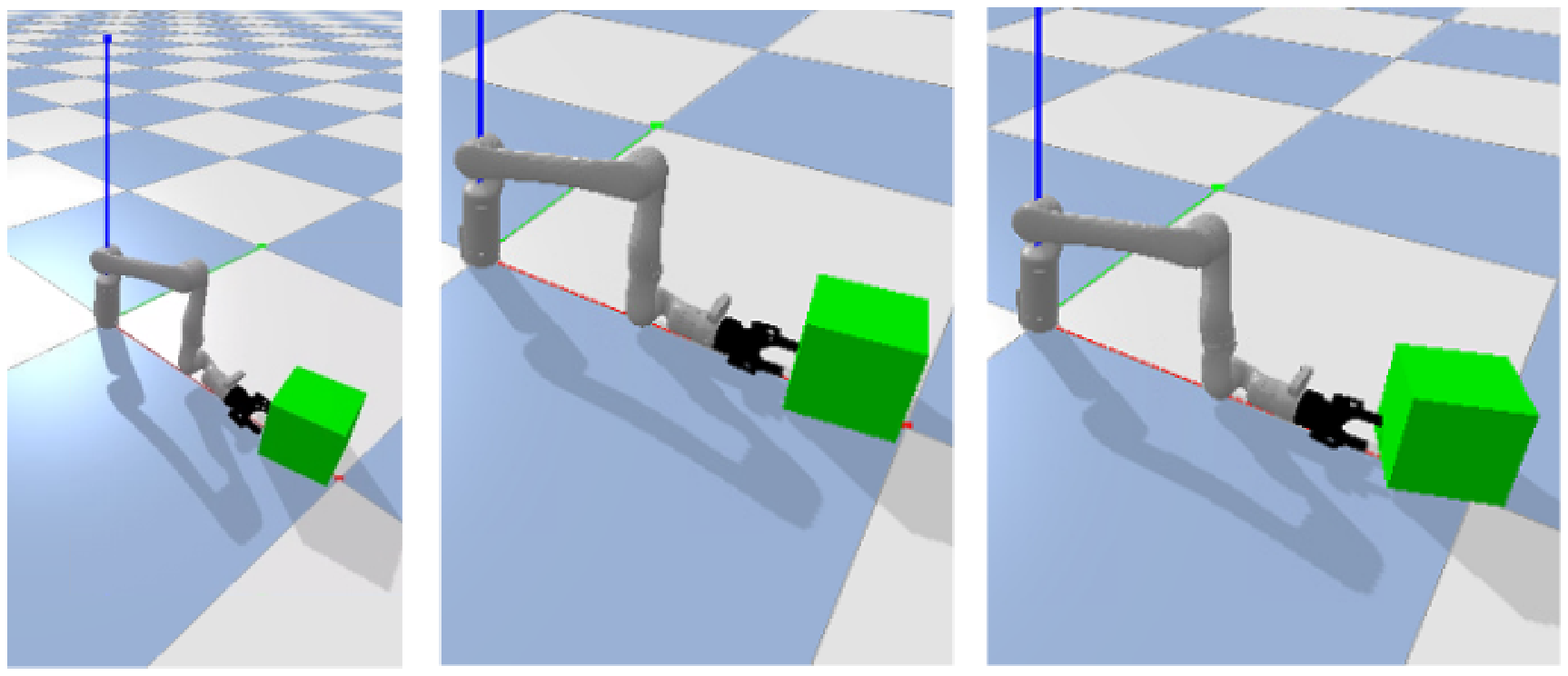}
         \caption{The OP task}
         \label{fig:push}
     \end{subfigure}
        \caption{\small{PyBullet simulation environments for the investigation of hardware utility effect on stochasticity.}}
        \label{fig:tasks}
\end{figure}

The intrinsic stochasticity in real-time simulation affects the trajectories and forces of the virtual robot models. To capture and measure the stochasticity, we repeatedly run the virtual robot for various robotic tasks in simulation and record its trajectories. Three different tasks are used in data collection, namely point-to-point reaching (P2P), pick-and-place (P\&P), and object-pushing (OP). These tasks have been chosen as simple and the most fundamental manipulation tasks that robotic arms are supposed to perform in any application \cite{yu2020meta}. With these three tasks, we want to gradually increase the computational burden on the system and evaluate the correlation with hardware usage in multiple scenarios. The procedures of the three tasks are illustrated in Fig.~\ref{fig:tasks}, with the following detailed interpretation.

\subsubsection{Point-to-Point Reaching (P2P)}
\label{sec:p2p}

This task involves moving the first three joints simultaneously from the robot's home position, pre-calibrated by the manufacturer, which is consistent between the virtual and the physical system. In this configuration, joint angles are at $\left[0, \frac{\pi}{12}, \frac{23\pi}{18}, 0, \frac{11\pi}{36}, \frac{\pi}{2}\right]$ rad. {In each iteration of moving the virtual or the physical robot for this experiment, we send constant $3 \times 10^{-3}\,deg$ position increment commands per each $1\,ms$ to the first three joints for a trivial motion.} Since zero joint acceleration inhibits the magnitude of actuator dynamics involved in the robot's behavior. After some preliminary tests, the joint speeds are calibrated so the same speed is applied to the real robot and simulation. For feedback, we continuously log all the signals mentioned in Sec.~\ref{sec:exp}, but we are mostly interested in the velocity ($V$) and torque ($T$) signals for they represent both kinematics and dynamics stochasticity. In the next section, we will investigate if the overall amount of variation in each of these signals has any significant correlation with average resource utility.

\subsubsection{Pick-and-Place Task (P\&P)}

This task follows the same logic, however, the starting position of the robot is defined so that the end-effector stands close to the green cube object in Fig.~\ref{fig:pnp}. The gripper grasps the object and then the second and third joint positions are moved by a constant speed of $0.3$ rad/s for the first 2 seconds. Then, the base joint position moves in a similar manner. Finally, for the last 2 seconds, the second and third joints move toward the target position, then let go of the object while opening the gripper. Here, again, we intended to keep dynamic complexity in the joint parameter space to a minimum. Instead, the complexity falls within the gripper function, the contact forces, and the dynamics of the object. The collected data is the same as before for this task.

\subsubsection{Object Pushing (OP)}

For the last experiment, a more complicated pushing task is designed. The manipulator starts from standing behind an object, with an open gripper. Then it starts moving in a straight horizontal line toward the object. The manipulator will keep on pushing after the collision while moving at the same speed and direction. Although this trajectory also seems simple in the task space, the translated 3DoF joint trajectories include accelerations and intricacies. Another detail that adds to the complexity, is the friction force between the object and the floor, which affects the joint torques through the contact force on the gripper. The collected data is the same as before for this task as well.

For each task, we run 50 trials and collect the robot output data including joint angles $\theta$, joint velocities $V$, joint torques $T$, and Cartesian position of the end-effector $(X, Y, Z)^T$. Each task runs for ten seconds (or equivalent simulation time) consisting of 10,000 steps at $1\,$kHz. The robot data is recorded at the same rate. We only activate three out of six joints of the robot model to simplify the simulation and reduce data requirements. To evaluate the consumption of the computer hardware resources, we also record the computer's real-time clock $t^c$ and the occupation of $CPU$, $GPU$, and $Memory$, inspired by the previous work on the evaluation of robot simulation platforms~\cite{ayala2020comparison, korber2021comparing, erez2015simulation}. We use Python library \emph{psutil} to collect the CPU and memory activity, and use Python module \emph{nvidia-smi} to collect the GPU utility. While running the simulations, no other tasks were active so the variations could be only explained by the running simulation. The analysis of the correlation between the stochastic robot motion and the hardware consumption is discussed in Sec~\ref{sec:result}.

\subsection{Experiment and Data Collection on A Physical Robot}

\begin{figure}[hbtp]
    \centering
    \includegraphics[width=0.35\textwidth]{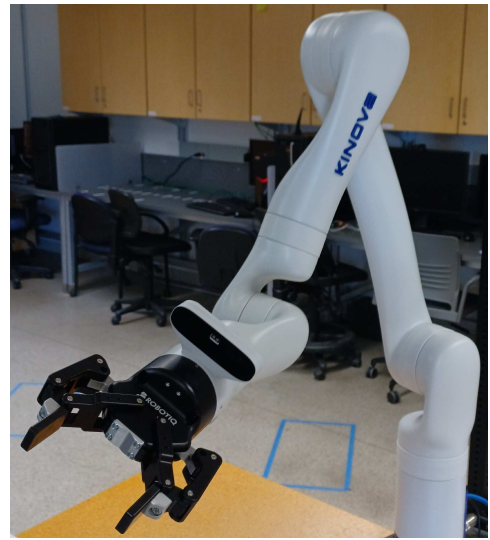}
    \caption{The physical experiment setup including the Kinova\textregistered~Gen3 robot and the Robotiq\textregistered~2F-85 gripper}
    \label{fig:kinova}
\end{figure}

In this paper, we also measure the RT-IS of a physical robot and compare it with that of the real-time simulation. For this purpose, we set up a Kinova\textregistered~Gen3 robot with a Robotiq\textregistered~2F-85 gripper, as in Fig.~\ref{fig:kinova}, corresponding to the simulation models in Sec.~\ref{sec:sim}. An off-the-shelf personal computer, equipped with Kinova\textregistered~Kortex\textsuperscript{TM}~Python API, transmits commands and feedback with the hardware at a rate of $1\,$kHz. {Noticeably, this is the highest rate supported by the feedback controller of physical hardware that indicates a response delay of less than $1\,$ms.} We assigned the P2P task, the simplest of the tasks mentioned in Sec.~\ref{sec:datasim}, to the physical robot to show that even the simplest real-time simulation is projecting sufficient stochasticity. The reason that we only choose P2P for the data collection on the physical robot is that P2P is sufficient to show the natural stochasticity of the physical robot. Also, P2P only involves the robot and no other external entities in the environment. This eliminates interference from any external dynamics, such as interactions with obstacles. The robot executes the same set of commands as the simulation configuration in Sec.~\ref{sec:datasim} while working in the manufacturer's generic position control mode. We continuously record the velocity ($V$) and torque ($T$) signals of the robot joints since they can sufficiently represent the stochasticity of robot dynamics. We use the data collected from the physical robot to compare it with its virtual counterpart, described in Sec.\ref{sec:sim}. The results will be presented in the following section.

\section{Correlation Analysis of RT-IS}
\label{sec:result}

Here, we analyze the collected data to assess the correlation between computer hardware and intrinsic stochasticity in real-time simulation. The results show that each of the three hardware benchmarks has a significant impact, depending on the task nature. Then, a comparable level of stochasticity is observed between the physical robot and its simulation. 
{
The overview of the analytical studies is illustrated in Fig.~\ref{fig:overview_ana}.
\begin{figure}[htbp]
\centering
\noindent

\begin{tikzpicture}[scale=1,font=\small]

\def\nw{2cm}
\def\nh{1cm}
\def\cn{0.15cm}

\definecolor{s_pink}{RGB}{255, 153, 153}
\definecolor{s_blue}{RGB}{153, 204, 255}
\definecolor{s_yellow}{RGB}{255, 230, 153}

\node[minimum height=\nh,minimum width=\nw, text width=\nw,align=center,draw, thick,fill=s_yellow, rounded corners=\cn] (sim_env) at (0cm,0cm) {Real-Time Simulation};
\node[minimum height=\nh,minimum width=\nw, text width=\nw,align=center,draw, thick,fill=s_blue, rounded corners=\cn] (sim_pi) at (0cm,1.5cm) {Heuristic Policy};

\draw[->,>=stealth,thick] (sim_pi.east) -- ([xshift=0.3cm] sim_pi.east) -- node[pos=0.5,align=right, anchor=east]{$\tilde{\tau}_t$} ([xshift=0.3cm] sim_env.east) -- (sim_env.east);
\draw[->,>=stealth,thick] (sim_env.west) -- ([xshift=-0.3cm] sim_env.west) -- node[pos=0.5,align=left, anchor=west]{$\tilde{v}_t$} ([xshift=-0.3cm] sim_pi.west) -- (sim_pi.west);

\draw[->,>=stealth,very thick,dashed] ([yshift=-0.5cm] sim_env.south) -- node[pos=0,align=left, anchor=north]{\footnotesize \textbf{RT-IS}}(sim_env.south);

\node[minimum height=\nh,minimum width=\nw, text width=\nw,align=center,draw, thick,fill=s_pink, rounded corners=\cn] (real_env) at (4cm,0cm) {Physical Robot};
\node[minimum height=\nh,minimum width=\nw, text width=\nw,align=center,draw, thick,fill=s_blue, rounded corners=\cn] (real_pi) at (4cm,1.5cm) {Heuristic Policy};

\draw[->,>=stealth,thick] (real_pi.east) -- ([xshift=0.3cm] real_pi.east) -- node[pos=0.5,align=right, anchor=east]{$\tau_t$} ([xshift=0.3cm] real_env.east) -- (real_env.east);
\draw[->,>=stealth,thick] (real_env.west) -- ([xshift=-0.3cm] real_env.west) -- node[pos=0.5,align=left, anchor=west]{$v_t$} ([xshift=-0.3cm] real_pi.west) -- (real_pi.west);

\draw[->,>=stealth,very thick,dashed] ([yshift=-0.5cm] real_env.south) -- node[pos=0,align=left, anchor=north]{\footnotesize \textbf{Physical Uncertainties}}(real_env.south);

\node[minimum height=0.1cm,minimum width=0.6cm,inner sep=0pt,draw,fill=black] (bar) at (1.3cm, 2.6cm) {};

\draw[->,>=stealth, thin] ([xshift=0.3cm] sim_pi.east) -- ([xshift=0.3cm, yshift=1.05cm] sim_pi.east);
\draw[->,>=stealth] ([xshift=-0.3cm] sim_pi.west) -- ([xshift=-0.3cm, yshift=0.8cm] sim_pi.west) -- ([xshift=2.3cm, yshift=0.8cm] sim_pi.west) -- ([xshift=2.3cm, yshift=1.05cm] sim_pi.west);
\draw[->,>=stealth] (bar.north) -- node[pos=0.55,align=left, anchor=west]{$\tilde{s}(t)$} ([yshift=0.7cm] bar.north);

\draw[->,>=stealth] ([yshift=-0.2cm] sim_env.west) -- ([xshift=-0.5cm, yshift=-0.2cm] sim_env.west) -- ([xshift=-0.5cm, yshift=1cm] sim_pi.west) -- node[pos=0.46,align=left, anchor=south]{\footnotesize CPU/GPU/Memory}([xshift=1.8cm, yshift=1cm] sim_pi.west) -- ([xshift=1.8cm, yshift=1.85cm] sim_pi.west);

\node[minimum height=0.5*\nh,minimum width=0.8*\nw, text width=0.8*\nw,align=center,draw, thick,fill=white, rounded corners=0.5*\cn] (pcc) at (1cm,3.6cm) {Correlation};

\node[minimum height=0.1cm,minimum width=0.6cm,inner sep=0pt,draw,fill=black] (bar1) at (5.3cm, 2.6cm) {};

\draw[->,>=stealth, thin] ([xshift=0.3cm] real_pi.east) -- ([xshift=0.3cm, yshift=1.05cm] real_pi.east);
\draw[->,>=stealth] ([xshift=-0.3cm] real_pi.west) -- ([xshift=-0.3cm, yshift=0.8cm] real_pi.west) -- ([xshift=2.3cm, yshift=0.8cm] real_pi.west) -- ([xshift=2.3cm, yshift=1.05cm] real_pi.west);
\draw[->,>=stealth] (bar1.north) -- node[pos=0.5,align=left, anchor=west]{$s(t)$} ([yshift=0.7cm] bar1.north);

\draw[->,>=stealth] ([yshift=0.15cm] bar.north) -- ([xshift=3.6cm, yshift=0.15cm] bar.north) -- ([xshift=3.6cm, yshift=0.7cm] bar.north);

\node[minimum height=0.5*\nh,minimum width=0.8*\nw, text width=0.8*\nw,align=center,draw, thick,fill=white, rounded corners=0.5*\cn] (cor) at (5cm,3.6cm) {Comparison};

\end{tikzpicture}
\caption{ The illustration of the analytical studies on (1) the correlation between the simulation data and the hardware resource consumption and (2) the comparison between the stochasticity of the real-time simulation and the physical robot.}
\label{fig:overview_ana}
\end{figure}
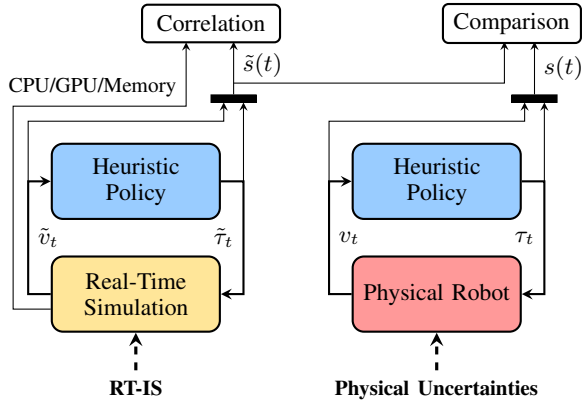

}

\subsection{Definition of Signals and Variables}

We define the outputs of the robot as signal $s(t)$ which includes the robot velocity and the torque values for the three active joints. Note that the $j$th repeat of the experiment signal, $s_j(t)$, differs from other repeats due to the RT-IS functionality in PyBullet. Thus, we define an average signal as,
\begin{equation}\label{eq:avg}
\bar{s}(t) = \frac{1}{N_{r}} \sum_{k=1}^{N_{r}} s_k(t),
\end{equation}
Where $N_r$ is the number of trials of the repeated signals. Then, the following deviation signal is defined to represent the extent of stochasticity,
\begin{equation}\label{eq:err}
\delta_j(t) = s_j(t) - \bar{s}(t).
\end{equation}
In addition to the average signal $\overline{s}(t)$, we also define the variability band of the signal $s$, using the standard deviation of the repeated signal values among different trials $s_j(t)$,
\begin{equation}\label{eq:sig}
\sigma(t) = \sqrt{\frac{1}{N_{r}} \sum_{j=1}^{N_{r}} (s_j(t) - \bar{s}(t))^2}.
\end{equation}
Also, we use $\Delta_j$ to represent the overall estimate of stochasticity for $j$th repeat of signal $s(t)$ using the following root-mean-square (RMS) function,
\begin{equation}\label{eq:rms}
\Delta_j = \sqrt{\frac{1}{N_{t}} \sum_{t=1}^{N_{t}} [\delta_j(t)]^2},
\end{equation}
Where $N_t$ is the number of timesteps of each signal $s_j(t)$. Besides, we calculate the average consumption of computer hardware $CPU$/$GPU$/$Memory$ for each trial $j$,
\begin{equation}
CPU_j = \frac{1}{N_{t}} \sum_{t=1}^{N_{t}} CPU_j(t),~GPU_j = \frac{1}{N_{t}} \sum_{t=1}^{N_{t}} GPU_j(t),
\end{equation}
\begin{equation}
Memory_j = \frac{1}{N_{t}} \sum_{t=1}^{N_{t}} Memory_j(t).
\end{equation}

\subsection{Stochasticity Correlation with Hardware Consumption }

Now, we have a measure $\Delta_j$ for the magnitude of the stochasticity in each trial $j$ for all the signals $s_j(t)$. Then, we use Pearson Correlation Coefficient (PCC) to analyze its correlation with the average hardware consumption $CPU_j$/$GPU_j$/$Memory_j$ for every trial $j$. For instance, the correlation between the second actuator's torque $\Delta_j$ in P\&P simulation and average $GPU$ percentage $GPU_j$ reads,
{
\begin{equation}\label{eg:pcc}
    r_{T_2,\,GPU}^{P\&P} = \frac{\sum_{j=1}^{N_{r}} (\Delta_j - \overline{\Delta})(GPU_j - \overline{GPU})}{\sqrt{\sum_{j=1}^{N_{r}} (\Delta_j - \overline{\Delta})^2} \sqrt{\sum_{j=1}^{N_{r}} (GPU_j - \overline{GPU})^2}},
\end{equation}
}
where,
\begin{equation}
    \overline{GPU} = \frac{1}{N_r} \sum_{j=1}^{N_r}GPU_j.
\end{equation}
{ By using the degree-of-freedom $n_{df} = N_r - 2$, and the following $t$-score,

\begin{equation}
    t_{T_2,\,GPU}^{P\&P} = \frac{r_{T_2,\,GPU}^{P\&P} \sqrt{n_{df}}}{\sqrt{1-(r_{T_2,\,GPU}^{P\&P})^2}},
\end{equation}

the $p$-value probability can be calculated as,

\begin{equation}
    p(t_{T_2,\,GPU}^{P\&P}, n_{df}) = \frac{\Gamma(\frac{n_{df}+1}{2})}{\sqrt{n_{df} \pi} \, \Gamma(\frac{n_{df}}{2})}\int_{-\infty}^{t_{T_2,\,GPU}^{P\&P}}{(1+\frac{x^2}{n_{df}})dx}
\end{equation}

with,

\begin{equation}
    \Gamma(z) = \int_0^{\infty}{u^{z-1}e^{-u}du}.
\end{equation}

The $p$-value expresses the probability of the hypothesis that a non-zero $r_{X,\,Y}$ could be in fact zero for a specific degree-of-freedom, considering the number of samples.} Thus, a $p$-value of less than $5\%$ shows a $95\%$ confidence in the calculated correlation coefficient.

The average $\Delta_j$ values for three tasks in the simulation are shown in Tab.~\ref{tab:avg}, which shows that the P2P task has the highest torque stochasticity, while P\&P has the highest velocity stochasticity.

\begin{table}[hbtp]
\caption{Average stochasticity ($\overline{\Delta}$) for velocity and torque signals in all three tasks}
\centering
\renewcommand{\arraystretch}{1.15}
\begin{tabular}{c c c c c}
\hline
\multirow{2}{4em}{Joint} & \multirow{2}{4em}{Signal} & \multicolumn{3}{c}{Task}\\ \cline{3-5}
  &  & \hfill P2P & P\&P & Push\\ \hline

\#1 & Velocity & 0.011 & 0.020 & 0.001\\ \cline{2-5}
  & Torque & 2.630 & 6.947 & 3.451\\ \hline

\#2 & Velocity & 0.011 & 0.029 & 0.007\\  \cline{2-5}
  & Torque & 8.631 & 5.818 & 2.756\\ \hline

\#3 & Velocity & 0.011 & 0.027 & 0.007\\  \cline{2-5}
  & Torque & 5.500 & 4.565 & 1.546\\ \hline
\end{tabular}
\label{tab:avg}
\end{table}

\begin{figure}[hbtp]
    \centering
    \includegraphics[width=0.48\textwidth]{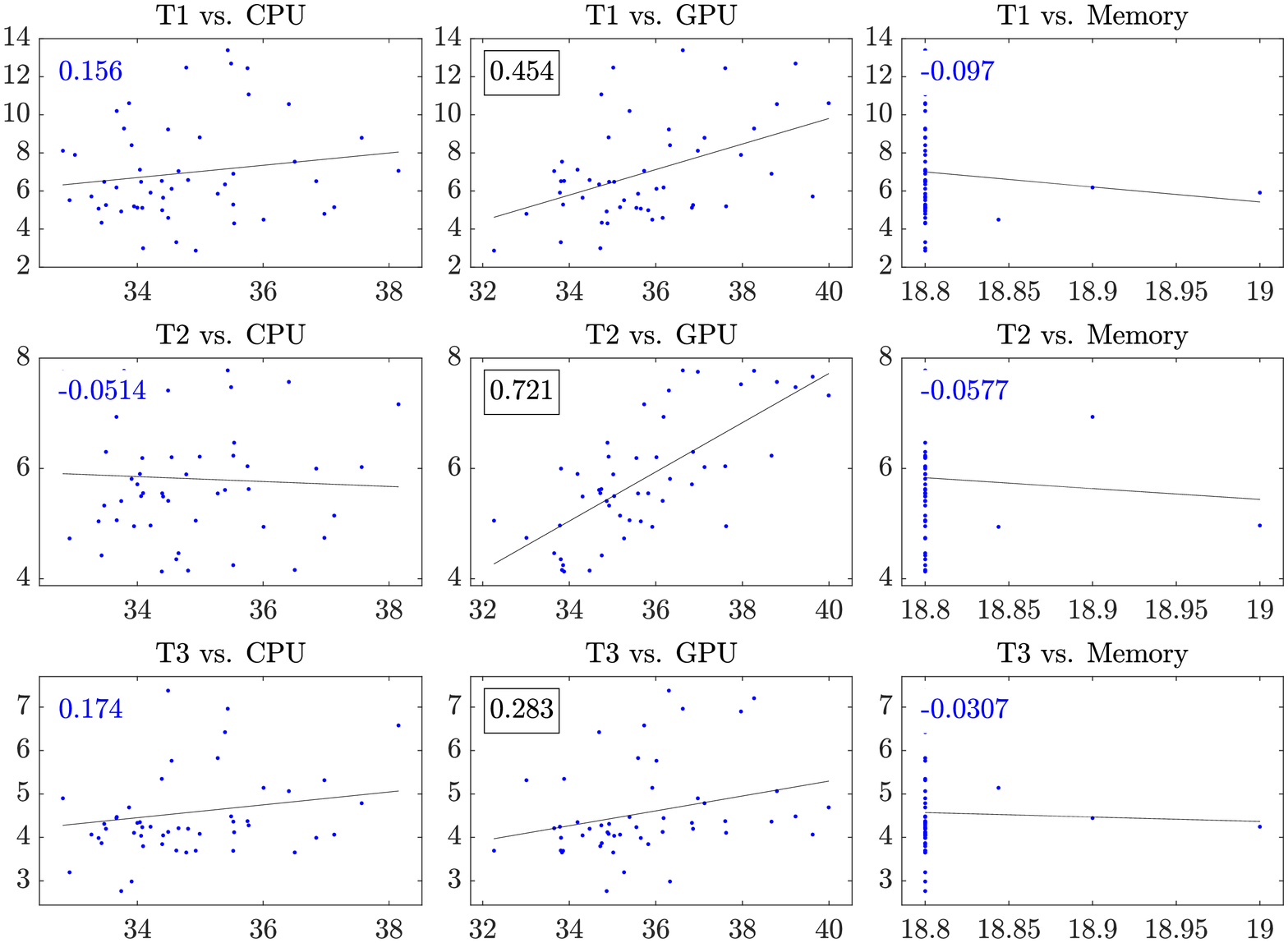}
    \caption{Correlation plots for RMS torque stochasticity vs. hardware consumption in P\&P simulation}
    \label{fig:corr}
\end{figure}

Also, for each simulation and signal, $\Delta_j$ is plotted against the average hardware consumption (e.g., $CPU_j$, $GPU_j$, or $Memory_j$) for that episode. Fig.~\ref{fig:corr} is a sample plot for torque stochasticity versus hardware utility consumption correlation in the P\&P trials. According to the PCC method and $p$-values, a significant positive effect of GPU usage on stochasticity is remarked (in the black text boxes). PCCs and $p$-value for the three tasks, P2P, P\&P, and OP, are respectively shown in Tab.~\ref{tab:p2p}, Tab.~\ref{tab:pnp}, and Tab.~\ref{tab:push}. The results show that, in P2P task, the only significantly important factor is CPU usage. A higher CPU percentage corresponds to more torque and less velocity stochasticity values for all actuators. As we previously pointed out, P2P has the highest torque stochasticity. This is a result of moving the two biggest links of the manipulator and exerting higher torque levels on the actuators. Larger torque demand applied to high-speed computational power, forces the CPU to cause a larger magnitude of stochasticity.

\begin{table}[hbtp]
\caption{PCCs (and $p$-values) for P2P simulation stochasticity vs. hardware benchmarks}
\centering
\renewcommand{\arraystretch}{1.15}
\begin{tabular}{c c c c c} \hline
Joint & Signal & CPU & GPU & Memory\\ \hline
\#1 & Velocity & \textbf{0.439} & -0.128 & 0.221\\
 &  & \textbf{(0.001)} & (0.375) & (0.123)\\ \cline{2-5}
 & Torque & \textbf{-0.447} & 0.252 & 0.118\\
 &  & \textbf{(0.001)} & (0.078) & (0.414)\\ \hline
 
\#2 & Velocity & \textbf{0.439} & -0.129 & 0.219\\
 &  & \textbf{(0.001)} & (0.373) & (0.126)\\ \cline{2-5}
 & Torque & \textbf{-0.394} & 0.245 & 0.250\\
 &  & \textbf{(0.005)} & (0.086) & (0.080)\\ \hline
 
\#3 & Velocity & \textbf{0.441} & -0.130 & 0.220\\
 &  & \textbf{(0.001)} & (0.370) & (0.125)\\ \cline{2-5}
 & Torque & \textbf{-0.433} & 0.243 & 0.231\\
 &  & \textbf{(0.002)} & (0.089) & (0.106)\\ \hline
\end{tabular}
\label{tab:p2p}
\end{table}

Unlike the previous examination, GPU is proven to be the only important factor in our designed P\&P scenario. Here, higher GPU usage is correlated with higher velocity and torque stochasticity in all three joints. In this scenario, which has the biggest velocity stochasticity among the three, motion is relatively faster and displacements, bigger. Therefore, the most influential factor in stochasticity is the GPU for its role in graphic visualization.

\begin{table}[hbtp]
\caption{PCCs (and $p$-values) for P\&P simulation stochasticity vs. hardware benchmarks}
\centering
\renewcommand{\arraystretch}{1.15}
\begin{tabular}{c c c c c} \hline
Joint & Signal & CPU & GPU & Memory\\ \hline
\#1 & Velocity & -0.250 & \textbf{0.546} & -0.105\\
 &  & (0.080) & \textbf{(0.000)} & (0.469)\\ \cline{2-5}
 & Torque & 0.156 & \textbf{0.454} & -0.097\\
 &  & (0.279) & \textbf{(0.001)} & (0.503)\\ \hline
 
\#2 & Velocity & -0.159 & \textbf{0.652} & -0.108\\
 &  & (0.269) & \textbf{(0.000)} & (0.457)\\ \cline{2-5}
 & Torque & -0.051 & \textbf{0.721} & -0.058\\
 &  & (0.723) & \textbf{(0.000)} & (0.690)\\ \hline
 
\#3 & Velocity & -0.157 & \textbf{0.627} & -0.114\\
 &  & (0.278) & \textbf{(0.000)} & (0.432)\\ \cline{2-5}
 & Torque & 0.174 & \textbf{0.283} & -0.031\\
 &  & (0.227) & \textbf{(0.047)} & (0.833)\\ \hline
\end{tabular}
\label{tab:pnp}
\end{table}

Finally, in the OP task, Memory usage overshadows other components (except for CPU impact on second actuator velocity) and shows significant relation with velocity stochasticity. This relatively complex scenario has in contrast the least amount of stochasticity compared to the other two. Yet, calculations in the OP task simultaneously consist of friction, contact force, object manipulation, and actuator dynamics. Therefore, the required calculation burden in each timestep is relatively bigger and consumes more $Memory$. Consequently, this factor would be the most responsible for stochasticity. It is also remarkable to see that only the velocity stochasticity is showing any significant relation to the hardware utility, due to the smaller forces and torques involved in this scenario.

To sum up, we can conclude that depending on the task nature, each hardware component or benchmark might be influential in determining the magnitude of added stochasticity to virtual simulations. $CPU$ tends to have the most important influence on stochasticity in the tasks with higher torques. $GPU$ is likely to be influential when higher velocity and faster visualization are involved in the task. Finally, for complex tasks with multiple dynamic elements, $Memory$ plays the main role in determining the stochasticity level.

\begin{table}[hbtp]
\caption{PCCs (and $p$-values) for OP simulation stochasticity vs. hardware benchmarks}
\centering
\renewcommand{\arraystretch}{1.15}
\begin{tabular}{c c c c c} \hline
Joint & Signal & CPU & GPU & Memory\\ \hline
\#1 & Velocity & 0.038 & -0.098 & \textbf{0.406}\\
 &  & (0.797) & (0.502) & \textbf{(0.004)}\\ \cline{2-5}
 & Torque & 0.087 & -0.130 & -0.007\\
 &  & (0.552) & (0.374) & (0.965)\\ \hline
 
\#2 & Velocity & \textbf{0.394} & -0.059 & \textbf{0.415}\\
 &  & \textbf{(0.005)} & (0.688) & \textbf{(0.003)}\\ \cline{2-5}
 & Torque & 0.109 & -0.103 & 0.013\\
 &  & (0.457) & (0.481) & (0.928)\\ \hline
 
\#3 & Velocity & 0.280 & -0.042 & \textbf{0.423}\\
 &  & (0.051) & (0.773) & \textbf{(0.002)}\\ \cline{2-5}
 & Torque & 0.125 & -0.107 & -0.002\\
 &  & (0.394) & (0.465) & (0.991)\\ \hline
\end{tabular}
\label{tab:push}
\end{table}

\subsection{Comparing the Stochasticity: Simulation vs. Real-world}

\begin{figure}[htbp]
    \centering
    \includegraphics[width=0.48\textwidth]{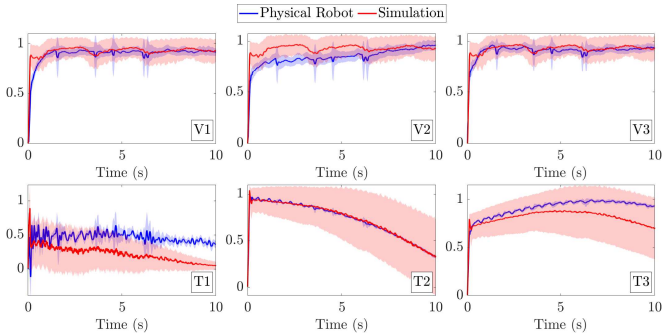}
    \caption{Velocity and torque signal (average and deviation) for P2P task}
    \label{fig:shade}
\end{figure}

\begin{figure}[htbp]
    \centering
    \includegraphics[width=0.48\textwidth]{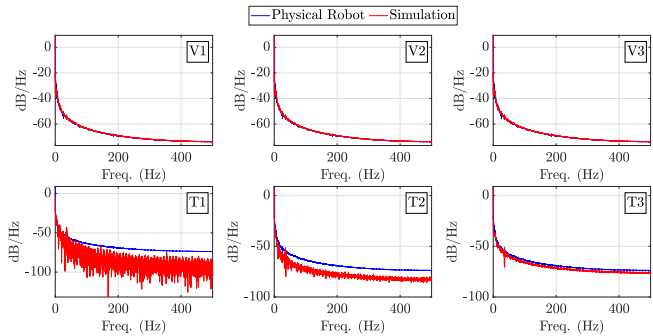}
    \caption{Power spectral density of velocity and torque signals for P2P task}
    \label{fig:psd}
\end{figure}

In the next step, we compare the stochasticity between the real-time simulation and the physical robot. Fig.~\ref{fig:shade} shows the normalized average velocity and torque of actuators, $\bar{s}(t)$, for both environments with solid lines. The shaded area corresponds to the variability range, $\sigma(t)$, of stochasticity. It is clear, particularly for torque signals, that simulation is noisier. Still, the power spectral density of all signals is quite similar (Fig.~\ref{fig:psd}) which indicates similar governing dynamic equations.

The RMS stochasticity value for all signals is also summarized in Tab.~\ref{tab:sim-real}. It emphasizes the higher stochasticity value in the simulation environment. Considering these results, the intrinsic stochastic functionality of real-time simulation software shows potential for improving simulation fidelity. If an RL agent learns to perform well in such stochastic simulation environments, its performance is less susceptible to deterioration when implemented on physical systems with intrinsic stochasticity~\cite{wang2020reinforcement, igl2019generalization}. This also proposes a measure of randomness for tuning the portion of domain randomization in increasing simulation stochasticity. In many cases, stochastic simulation environments like the real-time PyBullet simulation investigated in this paper may substitute heuristic domain randomization because of their trivial and unbiased nature.

\begin{table}[hbtp]
\caption{The RMS values of the stochasticity ($\overline{\Delta}$) in the P2P task (simulation and real-world)}
\centering
\renewcommand{\arraystretch}{1.15}
\begin{tabular}{c c c c}
\hline
Joint & Signal & Simulation & Real-World\\ \hline

\#1 & Velocity & 0.629 & 0.275\\ \cline{2-4}
  & Torque & 2.640 & 0.136\\ \hline

\#2 & Velocity & 0.629 & 0.263\\  \cline{2-4}
  & Torque & 8.592 & 0.245\\ \hline

\#3 & Velocity & 0.629 & 0.266\\  \cline{2-4}
  & Torque & 5.476 & 0.209\\ \hline
\end{tabular}
\label{tab:sim-real}
\end{table}

\section{Training and Test of RT-IS Powered Agents}\label{sec:exp}

In this section, we explain how the RT-IS is utilized to enhance the robustness of an RL agent. Firstly, an essential point-to-point (P2P) reaching robot task is formulated as a Markov Decision Process (MDP).

Then, we introduce how domain randomization and RT-IS are facilitated in the training process of an RL agent. Finally, we interpret the simulation and experimental configuration for the test studies.
{
The overview of the training and test process of the agents is illustrated in Fig.~\ref{fig:overall_train_test}.
\begin{figure}[htbp]
\centering
\noindent

\begin{tikzpicture}[scale=1,font=\small]

\def\nw{2cm}
\def\nh{1cm}
\def\cn{0.15cm}

\definecolor{s_pink}{RGB}{255, 153, 153}
\definecolor{s_blue}{RGB}{153, 204, 255}
\definecolor{s_yellow}{RGB}{255, 230, 153}

\node[minimum height=\nh,minimum width=\nw, text width=\nw,align=center,draw, thick,fill=s_yellow, rounded corners=\cn] (sim_env) at (0cm,0cm) {Real-Time Simulation};
\node[minimum height=\nh,minimum width=\nw, text width=\nw,align=center,draw, thick,fill=s_blue, rounded corners=\cn] (sim_pi) at (0cm,1.5cm) {Policy Training};

\draw[->,>=stealth,thick] (sim_pi.east) -- ([xshift=0.3cm] sim_pi.east) -- node[pos=0.5,align=right, anchor=east]{$\tilde{a}_t$} ([xshift=0.3cm] sim_env.east) -- (sim_env.east);
\draw[->,>=stealth,thick] (sim_env.west) -- ([xshift=-0.3cm] sim_env.west) -- node[pos=0.5,align=left, anchor=west]{$\tilde{s}_t$} ([xshift=-0.3cm] sim_pi.west) -- (sim_pi.west);

\draw[->,>=stealth,very thick,dashed] ([yshift=-0.5cm] sim_env.south) -- node[pos=0,align=left, anchor=north]{\footnotesize \textbf{KR/KR-RT-IS/KOR}}(sim_env.south);

\node[minimum height=\nh,minimum width=\nw, text width=\nw,align=center,draw, thick,fill=s_pink, rounded corners=\cn] (real_env) at (4cm,0cm) {Physical Robot};
\node[minimum height=\nh,minimum width=\nw, text width=\nw,align=center,draw, thick,fill=s_blue, rounded corners=\cn] (real_pi) at (4cm,1.5cm) {Policy Test};

\draw[->,>=stealth,thick] (real_pi.east) -- ([xshift=0.3cm] real_pi.east) -- node[pos=0.5,align=right, anchor=east]{$a_t$} ([xshift=0.3cm] real_env.east) -- (real_env.east);
\draw[->,>=stealth,thick] (real_env.west) -- ([xshift=-0.3cm] real_env.west) -- node[pos=0.5,align=left, anchor=west]{$s_t$} ([xshift=-0.3cm] real_pi.west) -- (real_pi.west);

\draw[->,>=stealth,very thick,dashed] ([yshift=-0.5cm] real_env.south) -- node[pos=0,align=left, anchor=north]{\footnotesize \textbf{Physical Uncertainties}}(real_env.south);

\end{tikzpicture}
\caption{ The illustration of the training and test of the agents, where $\tilde{s}_t$ and $\tilde{a}_t$ are the state and action of the agent in simulation, at certain time $t$, and $s_t$ and $a_t$ are those of the agent for the physical robot. Various RL agents are trained with whether the Kinematics Randomization (KR), the Kinematics Randomization Powered by RT-IS (KR-RT-IS), or the Kinematics-Observation Randomization (KOR) activated. The trained agents are directly tested on the physical robot affected by the physical uncertainties to evaluate their generalizability.}
\label{fig:overall_train_test}
\end{figure}
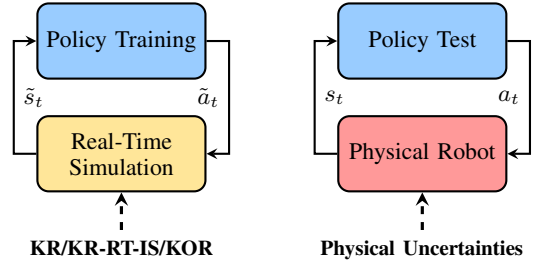
}

\subsection{Robot Task Formulation}

{The main objective of this paper is to provide RL methods for robotic manipulation tasks with a proposed technology for easy sim-to-real transfer.} It should be kept in mind that domain randomization is highly computationally expensive. Thus, we use a very simple but essential robot task, a P2P reaching task as the benchmark of this paper. A P2P reaching task requires the robot to start from an initial joint position and move until its end-effector reaches a desired Cartesian position. We formulate this robot task as an MDP $\mathcal{M} = (\mathcal{S},\mathcal{A}, f, R, \gamma)$, where $\mathcal{S}$ and $\mathcal{A}$ are respectively the state and action spaces of the robotic system, $f: \mathcal{S} \times \mathcal{A} \rightarrow \mathcal{S}$ represents the unknown state transition model of the system, $R: \mathcal{S} \times \mathcal{A} \rightarrow \mathbb{R}$ is the reward function determined according to the specific task, and $\gamma \in (0,1]$ is the discounting factor that balances between the current and the future rewards. The goal of the MDP problem is to solve the optimal policy $\pi:\mathcal{S} \rightarrow \mathcal{A}$ such that the accumulated reward is maximized.

For an $n$-degree-of-freedom (DoF) robot manipulator, the observation of $\mathcal{M}$, $s_t \in \mathcal{S}$, at time $t$, is defined as
\begin{equation}
    s_t = \left[ \, P^{\mathrm{e}}_t, \, \Theta_t, \, \sin(\Theta_t), \, \cos(\Theta_t), \, \mathbf{v}_t, \, P_{\mathrm{g}}, \, D \, \right],
\end{equation}
where $P^{\mathrm{e}}_t \in \mathbb{R}^3$ is the Cartesian position of the robot end-effector, $\Theta_t \in \mathbb{R}^n$ is the vector of the angles of the controllable robot joints, $\sin, \cos : \mathbb{R}^n \rightarrow \mathbb{R}^n$ are element-wise trigonometric functions of the robot joints, $\mathbf{v}_t \in \mathbb{R}^n$ is a vector of the joint velocities of the controllable joints, $P_{\mathrm{g}} \in \mathbb{R}^3$ is the Cartesian position of the goal, and $D \in \left\{ \, \mathrm{true}, \, \mathrm{false} \, \right\}$ is a binary signal indicating whether a trajectory has terminated. The action $a_t \in \mathcal{A}$ is selected as
\begin{equation}
    a_t = \mathbf{v}_t^{\mathrm{d}},
\end{equation}
where $\mathbf{v}_t^{\mathrm{d}} \in \mathbb{R}^n$ is the desired or commanded joint velocity {vector to be applied to the robot's active} joints at the current time step $t$. These observations provide enough relevant information for an agent to learn from as they all share some relationship that the agent can learn. We define the reward function as
\begin{equation}\label{eq:rew}
    {r_{t} = -R_1 E^{\mathrm{e}}_t - R_2 \|a_t\|^2}
\end{equation}
where $E^{\mathrm{e}}_t = \|P^{\mathrm{e}}_t - P_{\mathrm{g}}\|$ is the Euclidean distance between the end-effector and target positions, and $R_1, R_2 \in \mathbb{R}^+$ are gain hyper-parameters to be heuristically determined. In the reward function \eqref{eq:rew}, the term with the reaching error $E^{\mathrm{e}}_t$ is for the achievement of the P2P task. Also, we penalize the action $a_t$ to avoid dangerous actions that may be damaging to the real robot, such that the RL agent learns to move the robot at reasonable velocities and smoother trajectories. It is also worth noting that the relative value $R_1/R_2$, rather than the absolute values of the gains $R_1$ and $R_2$ are important. Finally, the terminal condition flag $D$ is determined as
\begin{equation}\label{eq:D}{
D = \left\{ \begin{array}{ll}
\mathrm{true}, & \mathrm{if} ~ t> N ~ \mathrm{or}~ E^{\mathrm{e}}_t < \varepsilon \\
\mathrm{false}, & \mathrm{otherwise}
\end{array}
\right.}
\end{equation}
where $N$ is a positive integer that indicates the maximum length of an episode and $\varepsilon \in \mathbb{R}^+$ is the threshold of the reaching error $E^{\mathrm{e}}_t$.

{
Our proposed approach to sim-to-real transfer can be modeled by this MDP formulation to explain the underlying theory. First, to better identify the differences in the formulation, we model two existing solutions for the sim-to-real problem. In \textit{domain adaptation}~\cite{gupta2017learning}, the stochasticity of the source MDP (the simulation environment) is formulated as a probability distribution $\mathcal{P}_{source}$ for the governing dynamics,
\begin{equation}
    s_{t+1} \sim \mathcal{P}_{source}(s_{t+1} \lvert s_t, a_t).
\end{equation}
In the same domain, a stochastic policy $\pi$ with the learned parameters $\theta_{source}$ is trained such that it maximizes the cumulative discounted reward,
\begin{equation}
    a_t \sim \pi(a_t \lvert s_t; \theta_{source}).
\end{equation}
After the transfer, a new distribution $\mathcal{P}_{target}$ is identified for the target environment dynamics (the physical system). Consequently, the policy parameters are mapped $\theta_{source} \mapsto \theta_{target}$ to compensate for the influence of the domain shift,
\begin{equation}
    \xRightarrow{\mathrm{Transfer}}
    \begin{cases}
        s_{t+1} \sim \mathcal{P}_{target}(s_{t+1} \lvert s_t, a_t), \\
        a_t \sim \pi(a_t \lvert s_t; \theta_{target}).
    \end{cases}
\end{equation}

Alternatively, in \textit{domain randomization}~\cite{pmlr-v87-muratore18a}, the dynamics $\mathcal{P}$ is modeled as a conditioned probabilistic function of the environment characteristics $\psi \in \mathbb{R}^{n_\psi}$ with $n_\psi$ parameters which may include inertia, structural parameters, friction coefficients and time delays in robotics,
\begin{equation}
    s_{t+1} \sim \mathcal{P}_{\psi}(s_{t+1} \lvert s_t, a_t, \psi).
\end{equation}
Note that the policy is also conditioned to the model characteristics,
\begin{equation}
    a_t \sim \pi_\psi(a_t \lvert s_t, \psi; \theta).
\end{equation}
$\psi$ is frequently sampled from $p(\psi)$ in between simulation experiences to increase the robustness of policy to multiple configurations of the model. If the target system's characteristics $\psi_{target}$ is sampled enough items during the training, the transfer would be straight-forward,
\begin{equation}
    \xRightarrow{\mathrm{Transfer}}
    \begin{cases}
        s_{t+1} \sim \mathcal{P}_{\psi_{target}}(s_{t+1} \lvert s_t, a_t, \psi_{target}), \\
        a_t \sim \pi_{\psi_{target}}(a_t \lvert s_t, \psi_{target}; \theta),
    \end{cases}
\end{equation}
otherwise, further training is necessary.

In the RT-IS approach to domain randomization, the dynamic calculations of the domain-randomized environment follow a stochastic time difference instead of the deterministic formulations. This timestep variability will affect all dynamic calculations consequently. We can model this by incorporating uncertainty to both state observation and action, or by conditioning the dynamics to $\tilde{\delta t} \sim p(\tilde{\delta t})$,
\begin{equation}
    s_{t+1} \sim \mathcal{P}(s_{t+\tilde{\delta t}} \lvert s_t, a_t, \tilde{\delta t}, \psi) \; \rightarrow \; s_{t+1} \sim \tilde{\mathcal{P}}(s_{t+1} \lvert \tilde{s}_t, \tilde{a}_t, \psi)
\end{equation}
A policy can be trained using the same stochastic signals,
\begin{equation}
    a_t \sim \pi(a_t \lvert s_t, \tilde{\delta t}, \psi; \theta) \; \rightarrow \; a_t \sim \pi(a_t \lvert \tilde{s}_t, \psi; \theta).
\end{equation}
After the transfer, time stochasticity still exists in the dynamics. However, it will follow a different distribution $\delta t \sim p(\delta t)$.
\begin{equation}
    \xRightarrow{\mathrm{Transfer}}
    \begin{cases}
        s_{t+1} \sim \mathcal{P}(s_{t+\delta t} \lvert s_t, a_t, \delta t, \psi_{target}), \\
        a_t \sim \pi(a_t \lvert s_t, \delta t, \psi_{target}; \theta).
    \end{cases}
\end{equation}

We argue that if the target stochasticity is present in the source distribution during training, the policy will be robust to the sim-to-real transfer.

}

\subsection{Configuration of Domain-Randomized Agents}\label{sec:dra}

Domain Randomization is the fundamental approach that we use to facilitate {robust sim-to-real for RL}. In brevity, domain randomization perturbs the simulation model of the environment and performs agent training on all incorporated models. {In this paper, we argue that utilizing the RT-IS can improve the feasibility and performance of straightforward domain-randomized RL agents. To verify this, we set up multiple RL baseline agents to conduct the studies. Our aim is to emphasize that even for basic baseline agents, enabling RT-IS can automatically improve the sim-to-real compatibility.}

\begin{itemize}

\item \textit{Kinematics-Randomized Agent (KRA):} 
A KRA is powered by the heuristic randomization in the robot kinematic model and trained in the non-real-time, or deterministic simulation mode. Firstly, we specify an imprecise kinematic model for the robot simulation to create the sim-to-real gap manually. Then, we sample the parameters of the kinematic model of the robot according to a specified distribution (randomization). The range of the randomization is heuristically determined. Finally, we train the KRA over all the randomized models using the domain-randomization algorithm~\cite{peng2018sim}. 

\item \textit{Kinematics Randomized Agent Powered by RT-IS (KRA-IS):} The configuration of KRA-IS is the same as the KRA, except that it is trained in the real-time simulation mode. With KRA-IS, we intend to show the advantage of RT-IS in the application of {sim-to-real in RL}.

\item \textit{Kinematics-Observation Randomized Agent (KORA):} Similar to the KRA, KORA also performs randomization on the kinematic model of the robot and is trained in the non-real-time simulation mode. Other than that, we manually add stochastic noise to the observation $s_t$ of the agent to impose randomization in the observation domain. The involvement of this agent is to verify whether a wider range of domain randomization contributes to better robustness.

\end{itemize}

Apart from the above-mentioned domain-randomized agents, we also configure the following non-randomized agents for comparison studies.

\begin{itemize}[leftmargin=*]

\item \textit{Non-randomized Agent with Precise Robot Kinematic Model (NA-P):} 
{The NA-P is a conventional Proximal Policy Optimization (PPO)~\cite{ppo} agent without any randomization}, trained in a non-real-time simulation environment using a precise robot kinematic model. This agent does not assume the existence of the sim-to-real gap, which serves as the most straightforward problem formulation for robot manipulation.

\item \textit{Non-randomized Agent with Imprecise Robot Kinematic Model (NA-I):}
The NA-I agent is a conventional RL agent without any randomization, trained in a non-real-time simulation environment using an imprecise robot kinematic model. The imprecise model is the same as the one used to train KRA. The NA-I admits the existence of the sim-to-real gap but attempts to resolve the issue conventionally. It indicates the performance of the conventional RL agents that are sensitive to the sim-to-real gap.

\end{itemize}

The PPO algorithm is used as the baseline of all agents since it is easy to implement, stable, and sample efficient. All agents are constructed as an actor-critic structure. Both the actor and critic are fully-connected forward neural networks with two hidden layers. The numbers of neurons in the first and the second layers are respectively 128 and 64. The training process of the agents is introduced in the following subsection.

\subsection{Agent Training}\label{sec:training}
 
The agents are trained in PyBullet~\cite{coumans2021} simulation environment which is an open-source platform that is compatible with the Open-AI Gym library and python RL libraries. PyBullet is an off-the-shelf and open-source simulation platform popularly used to simulate mechanical systems. It is powered by the Bullet physical engine and provides a powerful Python Interface that can be seamlessly called by the Open-AI Gym libraries. PyBullet provides a real-time simulation mode, enabled by the \textit{setRealTimeSimulation} method, that steps the simulation using the clock that depends on the operating system. The real-time simulation mode has not attracted much attention in RL since it brings uncertainties to the system's motion. However, believe that this mode is promising to enhance the robustness of conventional RL agents.

\begin{figure}[htbp]
    \centering
    \begin{subfigure}[b]{0.28\textwidth}
         \centering
         \includegraphics[height=5.5cm]{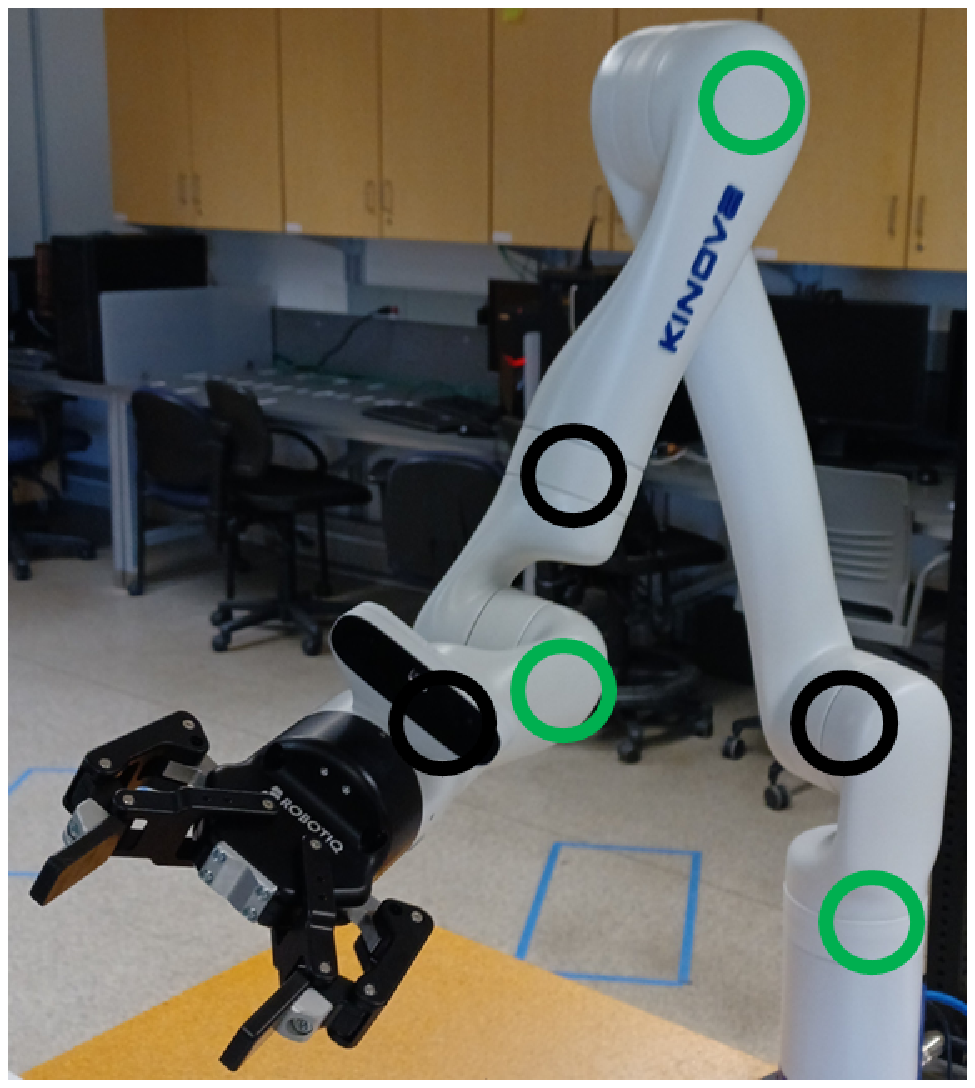}
         \caption{Real-robot}
         \label{fig:robotreal}
     \end{subfigure}
     \hfill
     \begin{subfigure}[b]{0.2\textwidth}
         \centering
         \includegraphics[height=5.5cm]{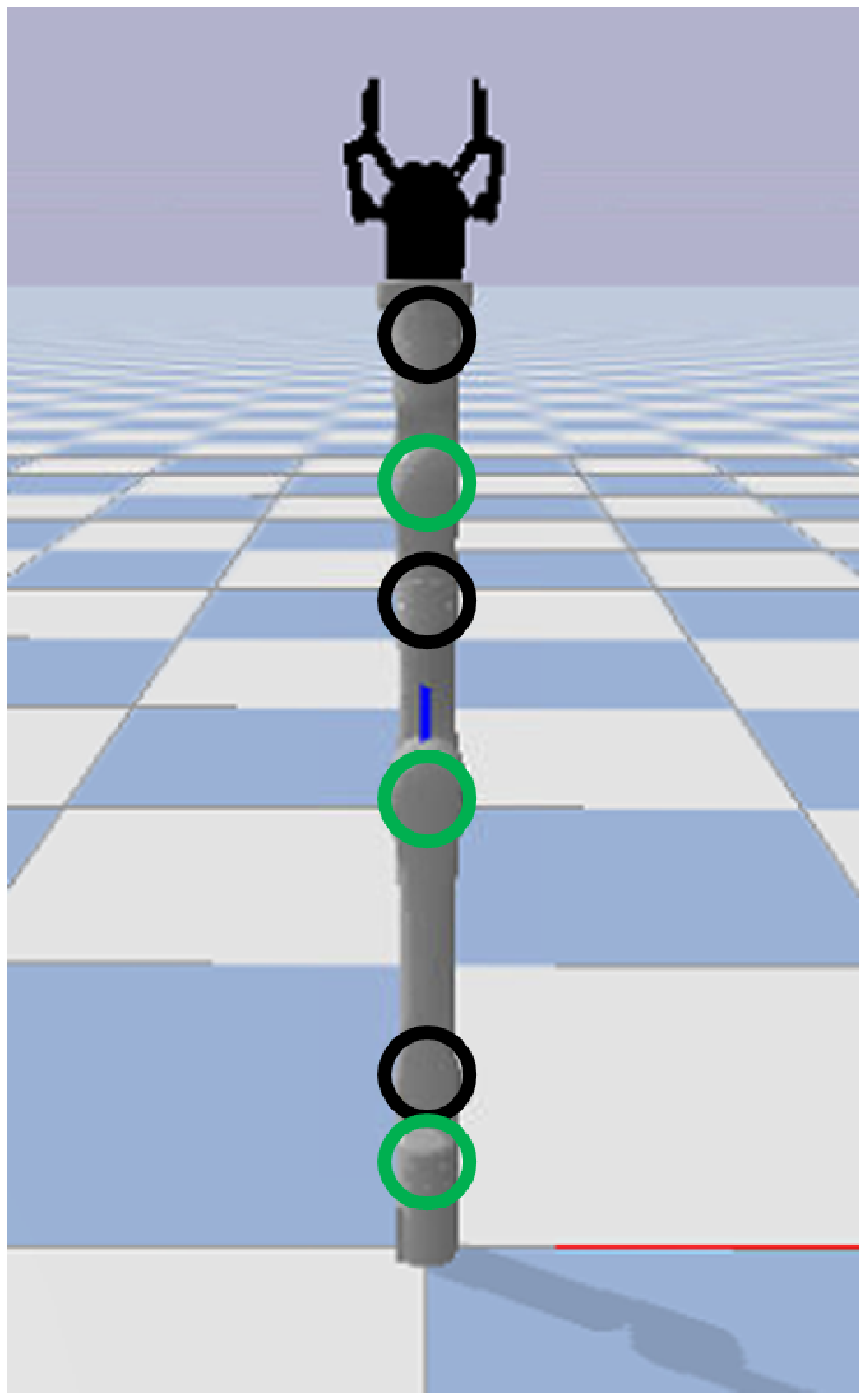}
         \caption{Simulated-robot}
         \label{fig:robotsim}
     \end{subfigure}
    \caption{The 6-DoF Kinova\textregistered~Robot and its PyBullet simulation model used for the experimental studies in this paper. The green and black circles are respectively representing the controllable joints and the manually-fixed joints.}
    \label{fig:robotcontrol}
\end{figure}

In this paper, we use a 6-DoF Kinova\textregistered~Gen3 robot, as shown in Fig.~\ref{fig:robotreal}, to conduct our experimental studies. The Kinova\textregistered~Gen3 lightweight robot has been widely applied to various research and engineering scenarios due to its extendability and versatility. The simulation model of the Kinova\textregistered~Gen3 robot in PyBullet is shown in Fig.~\ref{fig:robotcontrol} right. The kinematic parameters of the robot are obtained from Kinova\textregistered~official website~\cite{kinova}. 
{To simplify the problem and to emphasize the effectiveness of domain randomization, we only activate joints \#\,1, \#\,3, and \#\,5 of the robot as controllable, while manually fixing the remaining joints to zero position, as shown in Fig.~\ref{fig:robotcontrol}, to reduce the dimensions of the action and state spaces, which leads to a 3-DoF robot model.} Although the DoF of motion of the end effector is thus limited to three, it is sufficient for the P2P reaching task while ensuring the generalizability of the method.

For all training episodes, the base of the robot is spawned at the origin of the environment, $[\,0~0~0\,]^{\top}$, and the robot is initialized at the \textit{HOME} position, i.e., $\Theta_0 = 0$. The target is set at the Cartesian position $P_{\mathrm{g}}=[\,0.4~0.2~0.5\,]^{\top}$ measured in meters from the base of the robot. The KRA, KRA-IS, and KORA agents are trained in the following manner.

\begin{itemize}[leftmargin=*]

\item The KRA and the KRA-IS are both trained by randomizing the domain of the kinematic model. Specifically, the length of both the first and second links of the robot arm is sampled from uniform distribution among $100\% \pm 1\%$ of the original model. The range of the kinematic model randomization is kept small but sufficient to cover the possible difference between the simulation environment and the real world.

\item The KORA is subject to the same kinematic model randomization as KRA and KRA-IS. In addition, its observation domain is also randomized with noisy signals added. The amplitude of the noisy signal is determined as no more than 5\% of the robot joint velocity.

\item The NA-I is trained using an imprecise robot kinematic model in simulation. Compared to the original robot kinematic model, the first link is 0.04\% shorter and the second link is 0.04\% longer. The NA-P is trained using the precise kinematic model of the robot arm.

\end{itemize}

{The hyper-parameters of the reward \eqref{eq:rew} are assigned as $R_1 = 2 \times 10^{-5}$, $R_2 = 10^{-6}$.} The threshold of the reaching error is determined as $\varepsilon = 0.05\,$m. The training for all agents was run for 1000 epochs. For each agent, the policy is updated after each epoch which contains 5 episodes. Each episode contains 200 steps, i.e., $N=200$. {The discrete sampling time of the training process is $25\,ms$ which is sufficiently bigger than the actuator delay of $1\,ms$}. The configuration of the test study is interpreted in the next subsection. The training results are discussed and analyzed in Sec.~\ref{sec:exp_result}.

\subsection{Test Studies}\label{sec:testing}

Before implementing the trained agents on the real-world robot, test studies are needed to verify the generalizability and applicability of the agents. There are two main reasons for the test studies, firstly, to ensure that the agents perform the task correctly, and secondly, to test and compare the performance of the trained agents. By testing in simulation, we can ensure that the agents learned to execute the task safely and reliably. Moreover, the test performance of the agents gives us insight into how they may perform in the real world. 

\begin{figure}[htbp]
     \centering
     \begin{subfigure}[b]{0.11\textwidth}
         \centering
         \includegraphics[height=2.3cm]{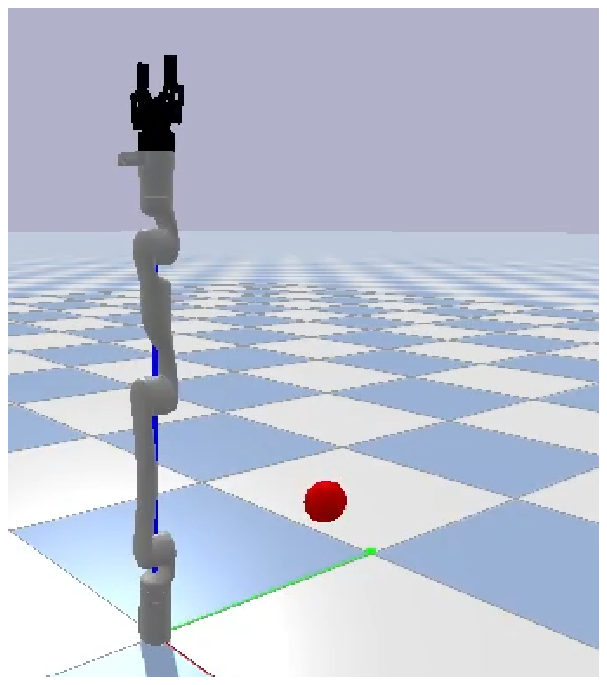}
         \caption{Initial}
         \label{fig:init_V_1_sim}
     \end{subfigure}
     \hfill
     \begin{subfigure}[b]{0.11\textwidth}
         \centering
         \includegraphics[height=2.3cm]{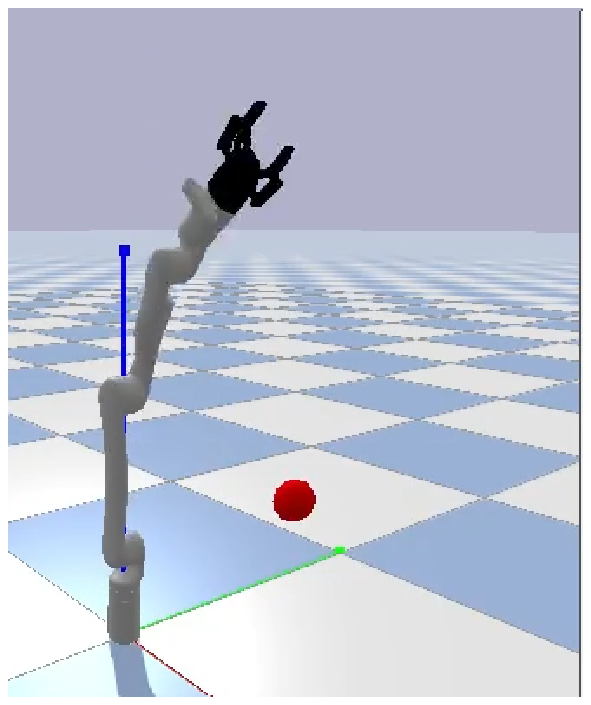}
         \caption{Moving}
         \label{fig:init_V_2_sim}
     \end{subfigure}
     \hfill
     \begin{subfigure}[b]{0.11\textwidth}
         \centering
         \includegraphics[height=2.3cm]{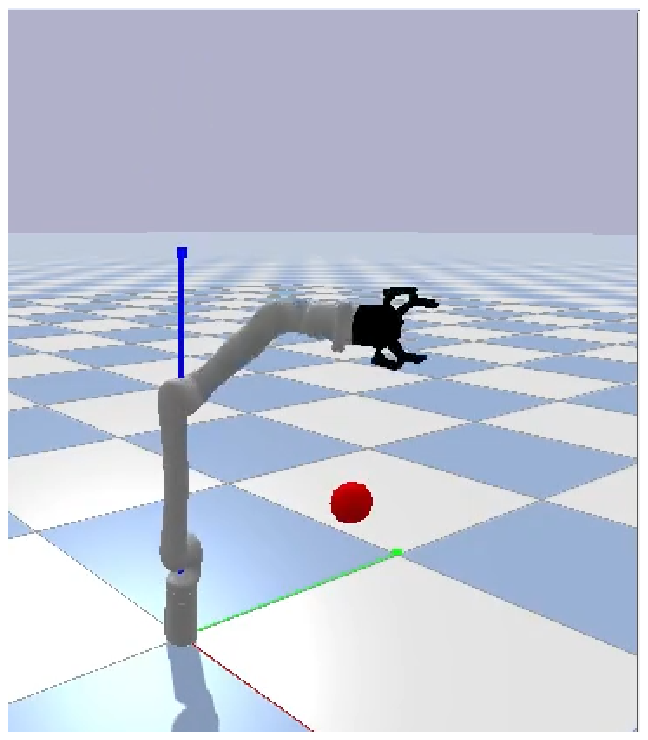}
         \caption{Moving}
         \label{fig:init_V_3_sim}
     \end{subfigure}
     \hfill
     \begin{subfigure}[b]{0.11\textwidth}
         \centering
         \includegraphics[height=2.3cm]{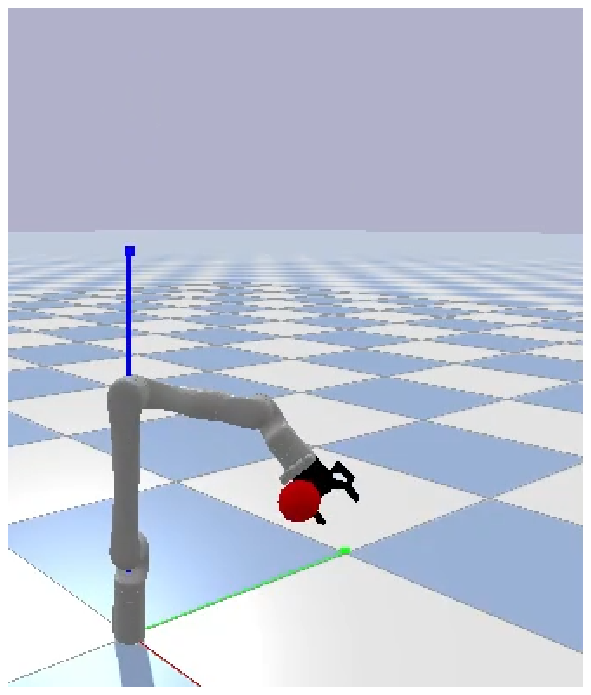}
         \caption{Target}
         \label{fig:init_Cost_1_sim}
     \end{subfigure}
     \hfill
     \begin{subfigure}[b]{0.11\textwidth}
         \centering
         \includegraphics[height=2.3cm]{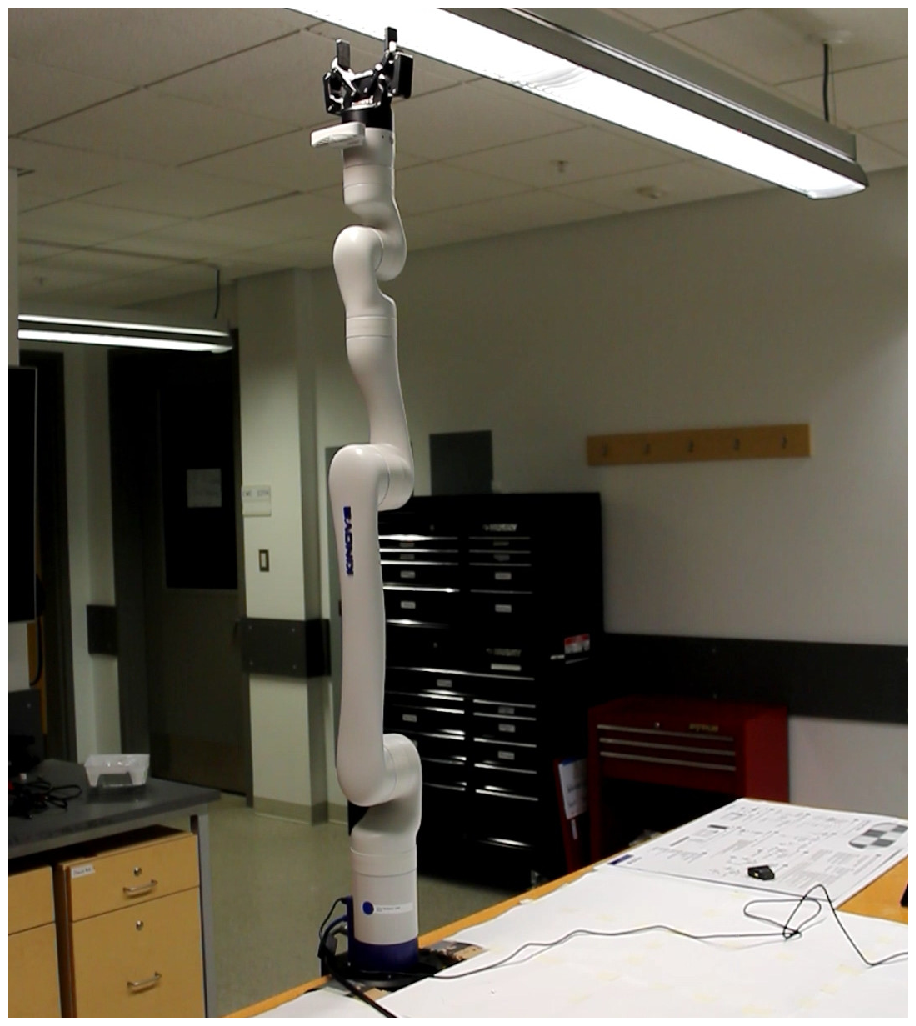}
         \caption{Initial}
         \label{fig:init_V_1}
     \end{subfigure}
     \hfill
     \begin{subfigure}[b]{0.11\textwidth}
         \centering
         \includegraphics[height=2.3cm]{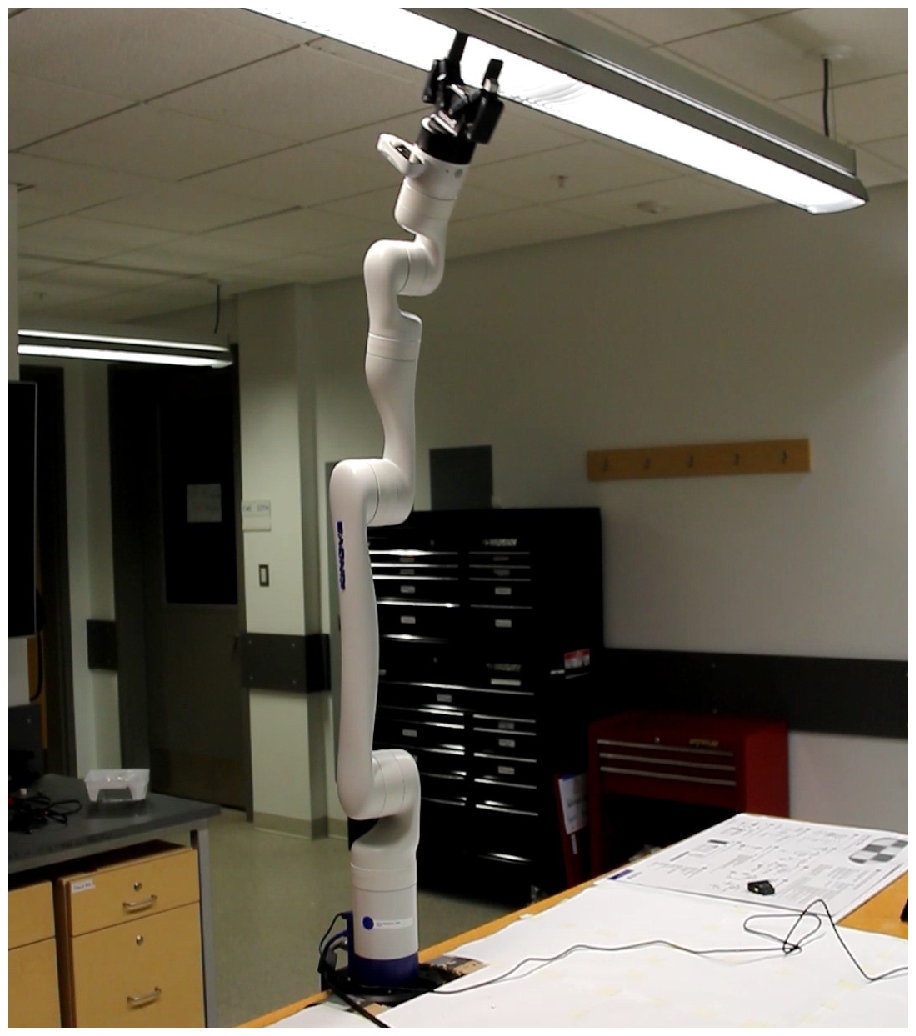}
         \caption{Moving}
         \label{fig:init_V_2}
     \end{subfigure}
     \hfill
     \begin{subfigure}[b]{0.11\textwidth}
         \centering
         \includegraphics[height=2.3cm]{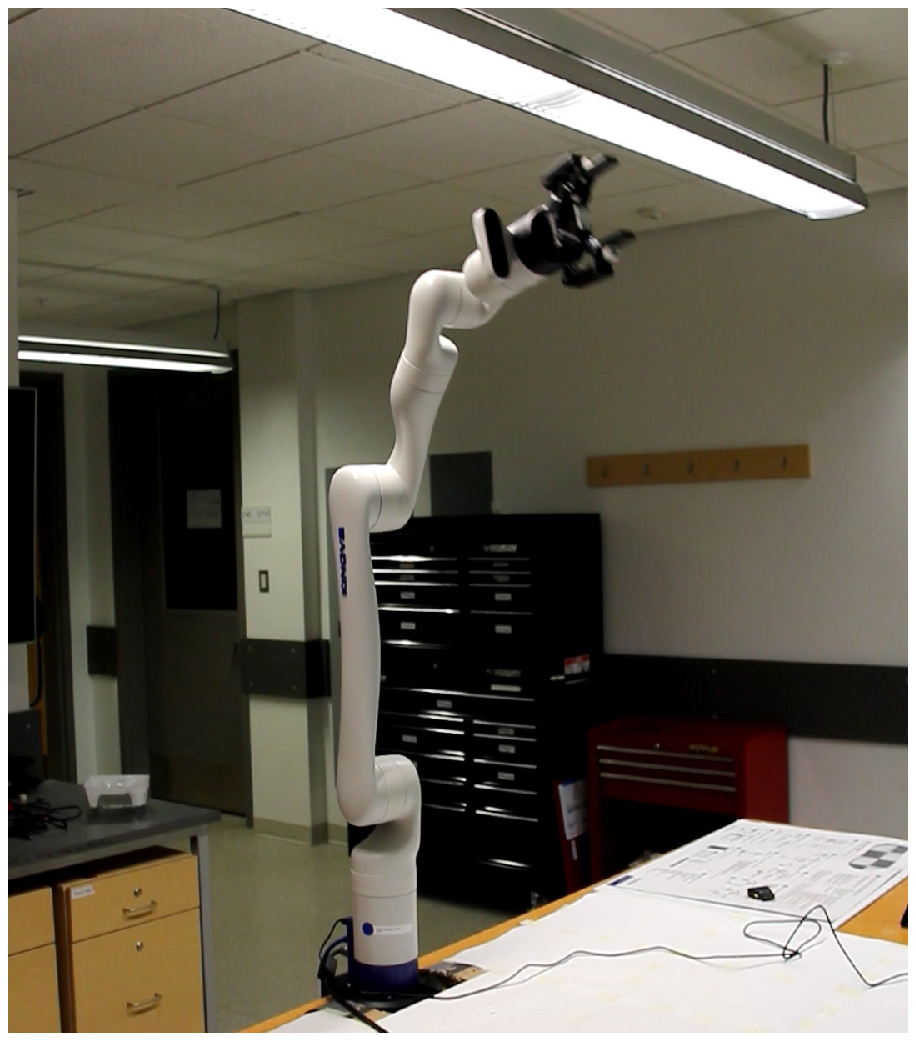}
         \caption{Moving}
         \label{fig:init_V_3}
     \end{subfigure}
     \hfill
     \begin{subfigure}[b]{0.11\textwidth}
         \centering
         \includegraphics[height=2.3cm]{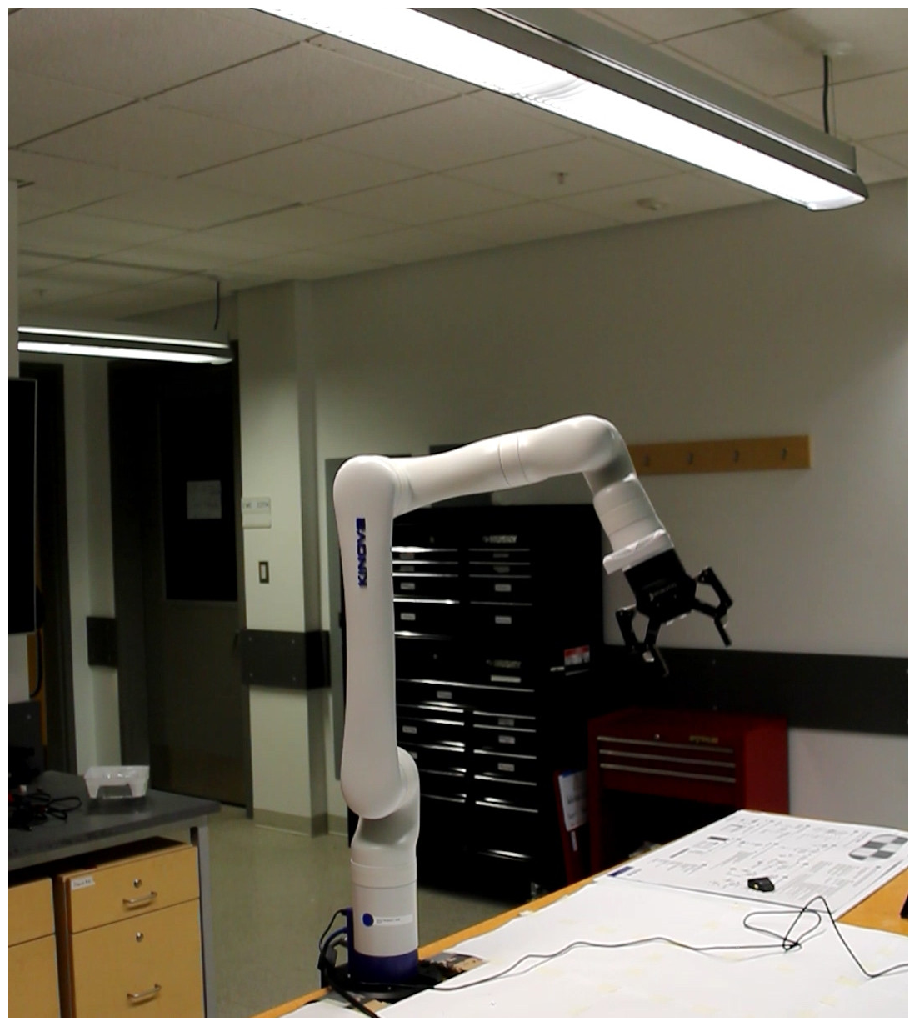}
         \caption{Target}
         \label{fig:init_Cost_1}
     \end{subfigure}
        \caption{The motion of the robot in simulation (a)-(d) and hardware (c)-(h) test studies. The figures respectively show the robot starting from the initial position ((a) and (e)), moving towards the target position ((b), (c) and (f), (g)), and the robot reaching the target position ((d) and (h)).}
        \label{fig:exp_demo}
\end{figure}

Both tests in the PyBullet simulation and on the real robot platform are conducted. In all test studies, the robot starts from zero joint position. The simulation test is conducted in the non-real-time mode to avoid simulation uncertainties. To perform reasonable result analysis, we run 500 test trials in simulation. For the hardware test, 50 test trials are recorded for each agent. The examples of test trials in simulation and on the robot hardware are presented in Fig.~\ref{fig:exp_demo} The Analysis and discussion of the test results are presented in Sec.~\ref{sec:exp_result}.

\section{Result Discussion}\label{sec:exp_result}

In this section, we evaluate the training and test results of the agents introduced in Sec.~\ref{sec:dra}. Firstly, the training performance of the agents is presented based on their learning curves and the associated numerical metrics. Then, the results of the agents in both simulation and hardware experiment test studies are discussed.

\subsection{The Training Performance}

\begin{figure}[htbp]
    \centering
    \includegraphics[width=0.5\textwidth]{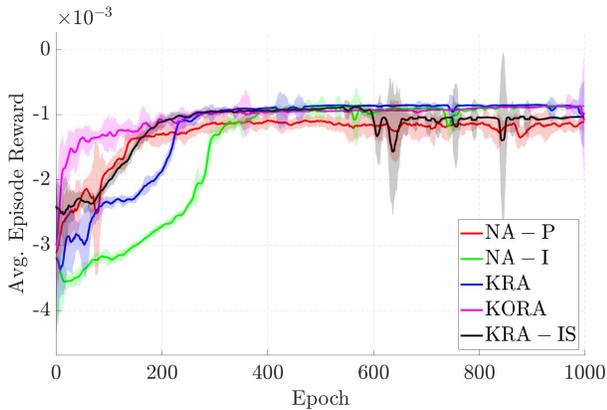}
    \caption{Training results of all agents (Average Episode Reward vs. Epoch for three random seeds per agent)}
    \label{fig:traincurves}
\end{figure}

We use the learning curves of the agent as the episode increases as the main metrics to evaluate the agent training, as shown in Fig.~\ref{fig:traincurves}. Based on this, Four additional numerical metrics as follows are also used.
\begin{itemize}[leftmargin=*]
\item $\mathcal{R}_{\mathrm{ini}}$: the initial reward calculated by averaging the instant reward after the initial 10\% of epochs, which shows the performance of the initial policies;
\item \textbf{$\mathcal{R}_{\mathrm{ult}}$}: the ultimate reward calculated by averaging the instant reward over the final 10\% of epochs, which depicts the performance of the trained policies; 
\item \textbf{$\mathcal{T}_{\mathrm{hlf}}$}: the half-trained time, a percentage of the episode at which the accumulated reward first exceeds 50\% of $\mathcal{R}_{\mathrm{ult}}$
\item {\textbf{$r_{time}$}: the ratio of the training time (based on the wall clock time) to the training time of the NA-P agent}
\end{itemize}

The values of the numerical metrics in the training process are shown in Tab.~\ref{tab:num_metrics}. The main phenomenon we notice in Fig.~\ref{fig:traincurves} is how domain randomization affects the training of RL agents. More noise is witnessed on the learning curves of the domain-randomization agents, including KRA, KRA-IS, and KORA. This is because all these agents are trained over various robot and environment models. The attempt to perform training over multiple environmental models means the sacrifice of the stability of the training process. Also, the KRA-IS, dominated by not only the manual model randomization but also the RT-IS, achieves the lowest ultimate reward, although the inferiority is insignificant. This is reasonable to understand since it involves the highest extent of randomization over other RL agents. Similar information can also be inferred from  Tab.~\ref{tab:num_metrics} that domain randomization agents KRA-IS, KRA, and KORA have lower ultimate reward $\mathcal{R}_{\mathrm{ult}}$ and reward increment $\mathcal{R}_{\mathrm{inc}}$. Besides, the learning curves of these agents show a slower convergence rate than NA-I and NA-P, which is also reflected by larger values of $\mathcal{T}_{\mathrm{hlf}}$ metric in Tab.~\ref{tab:num_metrics}. Overall, domain randomized agents tend to have inferior training performance than the non-randomized ones due to the sacrifice for generalization. In the next subsection, we will evaluate whether the performance tradeoff paid off for the test studies.

\linespread{1.2}
\begin{table}[htbp]
\caption{Numerical Metrics of the Agent Training}
    \centering
    \begin{tabular}{c|c|c|c|c}
    \hline
        \textbf{Agent} & {$\mathcal{R}_{\mathrm{ini}}\times10^3$} & {\textbf{$\mathcal{R}_{\mathrm{ult}}\times10^3$}} & {$\mathcal{T}_{\mathrm{hlf}}$} &
        {$r_{time}$} \\
    \hline
        KRA-IS & $-2.33$ & $-1.21$ & 10.2\% & 2.03\\
    KRA & $-2.71$ & $-0.85$ & 21.5\% & 1.57\\
        KORA & $-3.42$ & $-0.91$ & 26.9\% & 1.54\\
        NA-P & $-2.32$ & $-1.09$ & 12.4\% & 1.00\\
        NA-I & $-1.43$ & $-0.84$ & 9.3\% & 1.54\\
    \hline
    \end{tabular}
    \label{tab:num_metrics}
\end{table}
\linespread{1}

\begin{figure*}[htbp]
    \centering
    \begin{subfigure}[b]{0.19\textwidth}
         \centering
         \includegraphics[width=0.98\textwidth,clip]{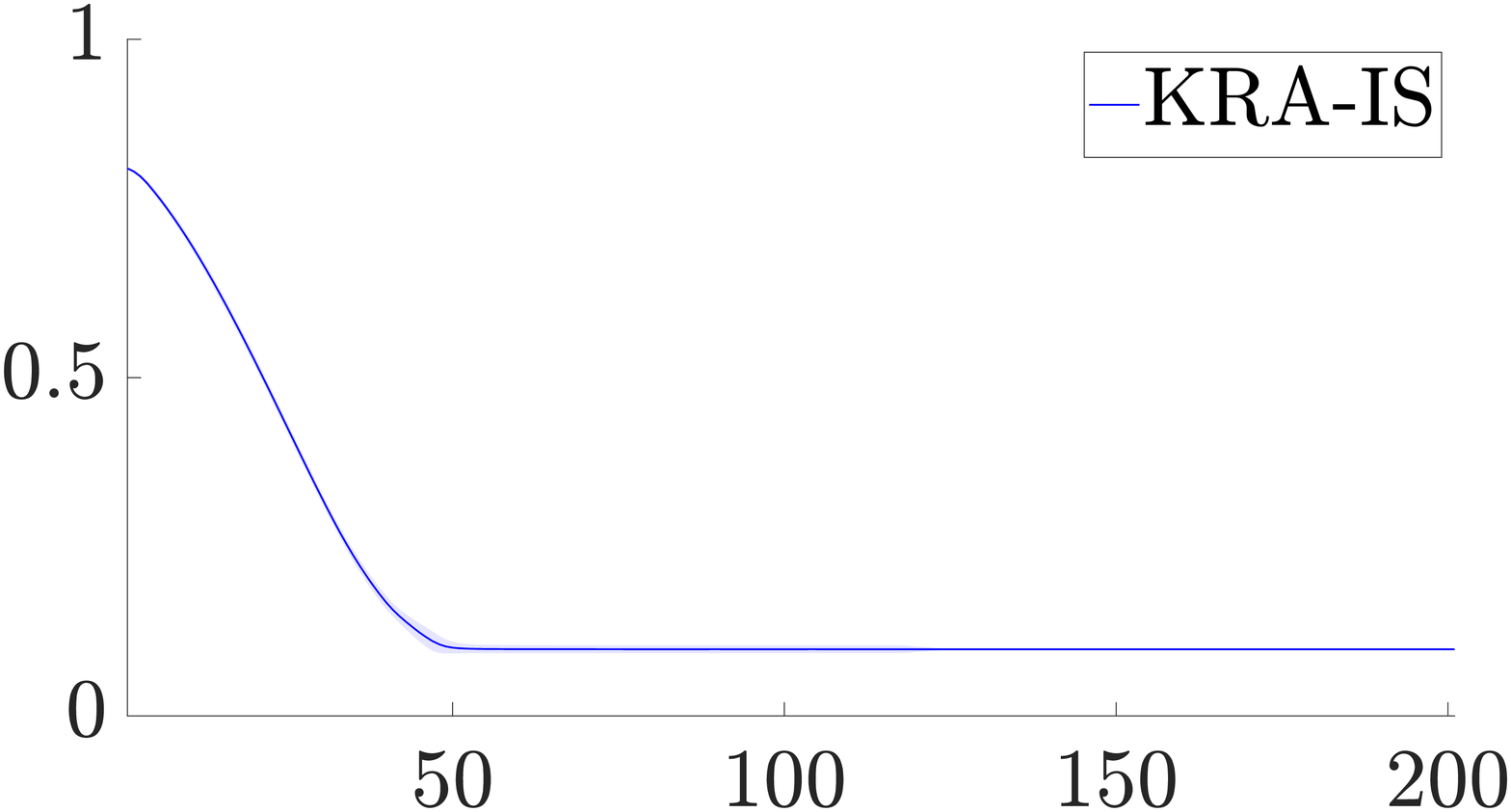}
    \end{subfigure}
        \begin{subfigure}[b]{0.19\textwidth}
         \centering
         \includegraphics[width=0.98\textwidth,clip]{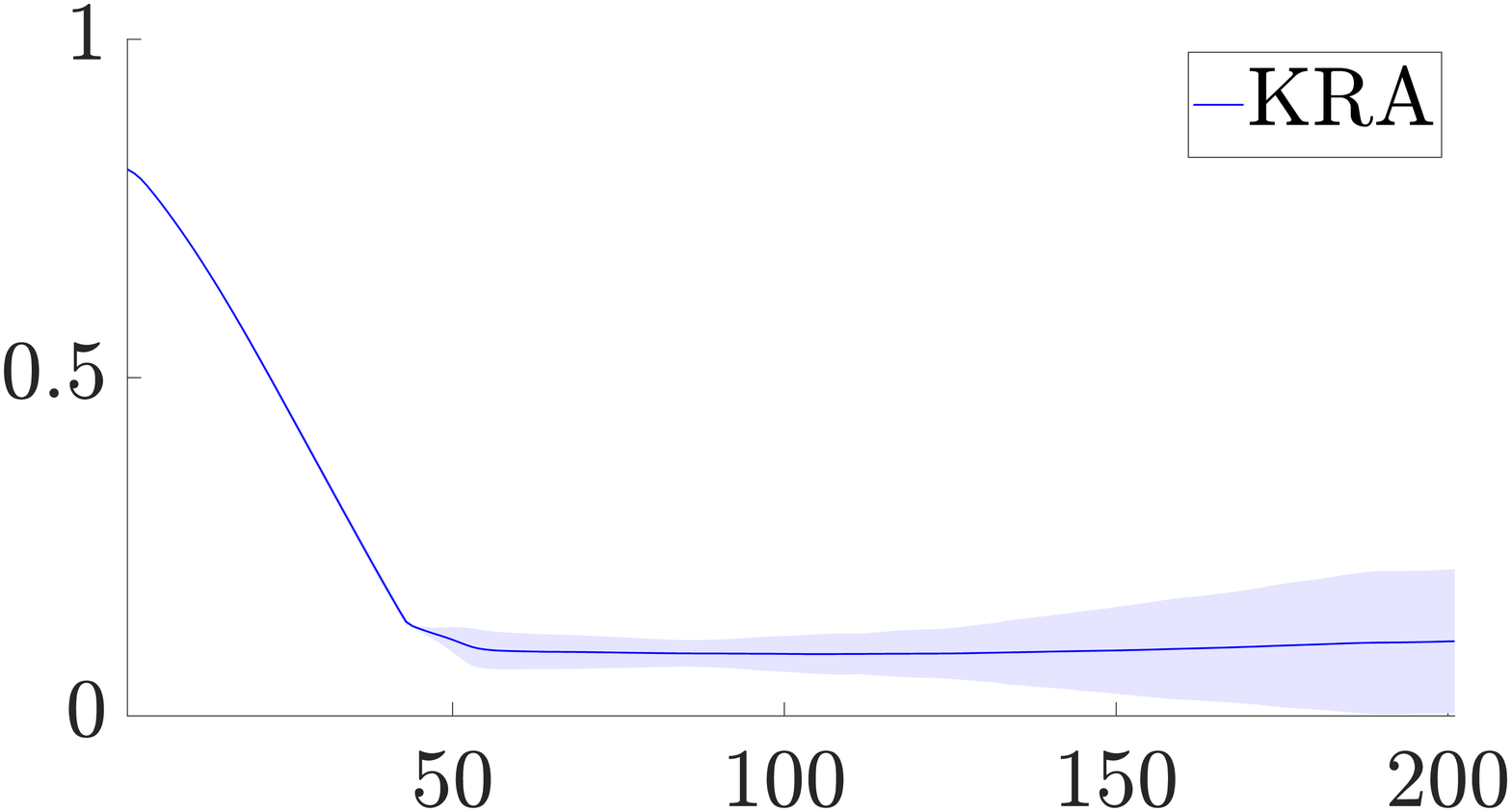}
    \end{subfigure}
    \begin{subfigure}[b]{0.19\textwidth}
         \centering
         \includegraphics[width=0.98\textwidth,clip]{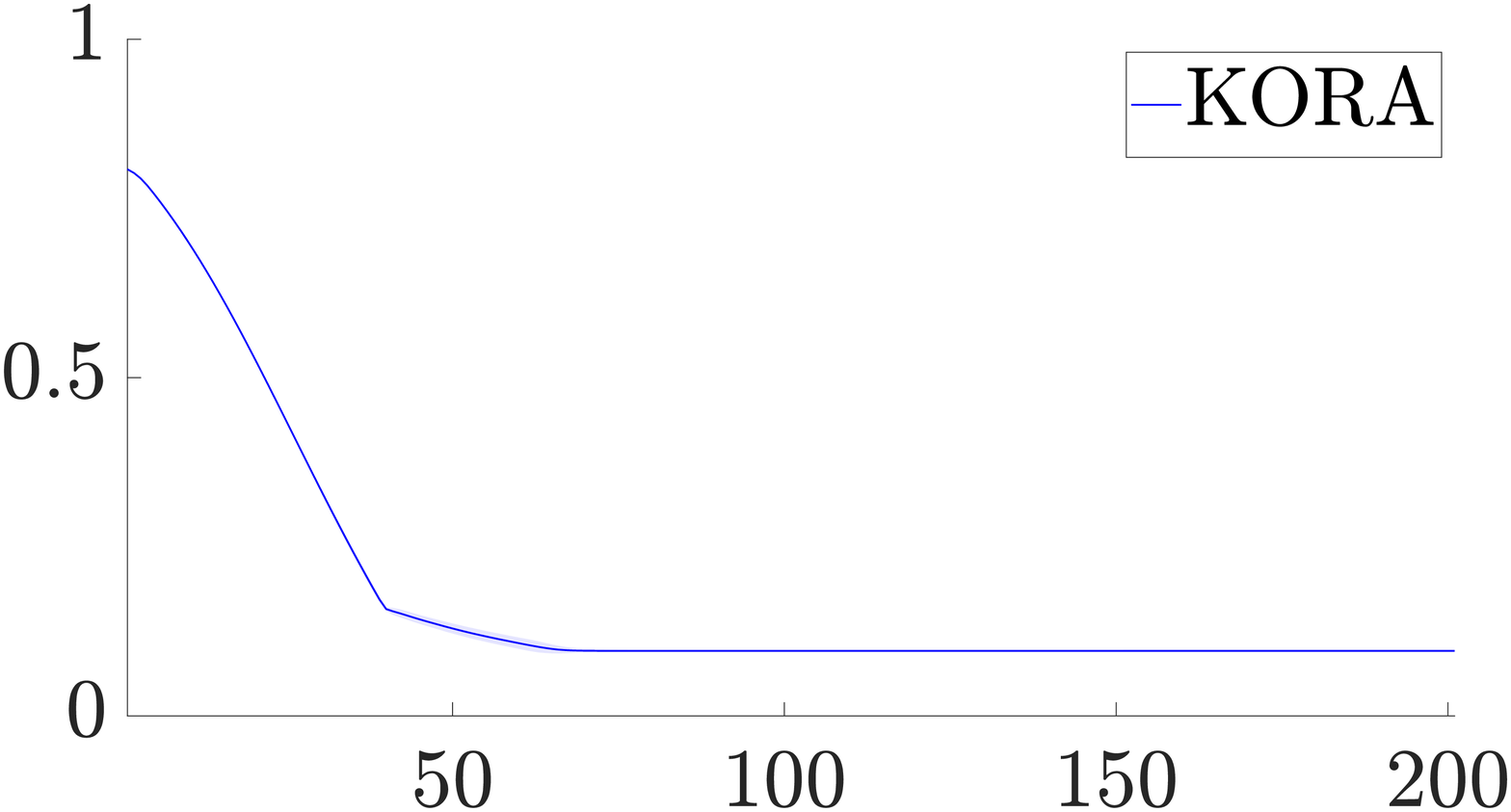}
    \end{subfigure}
    \begin{subfigure}[b]{0.19\textwidth}
         \centering
         \includegraphics[width=0.98\textwidth,clip]{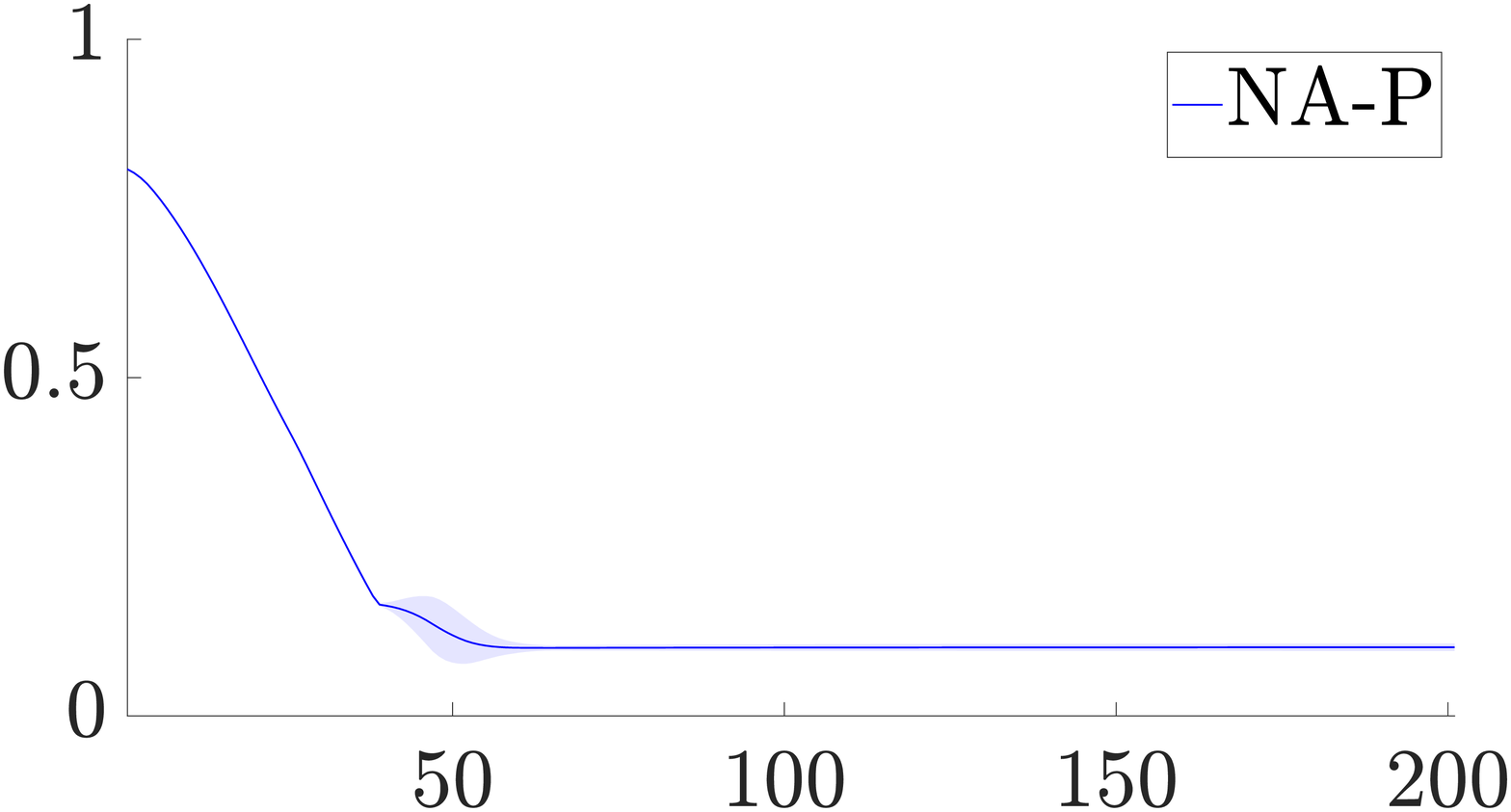}
    \end{subfigure}
    \begin{subfigure}[b]{0.19\textwidth}
         \centering
         \includegraphics[width=0.98\textwidth,clip]{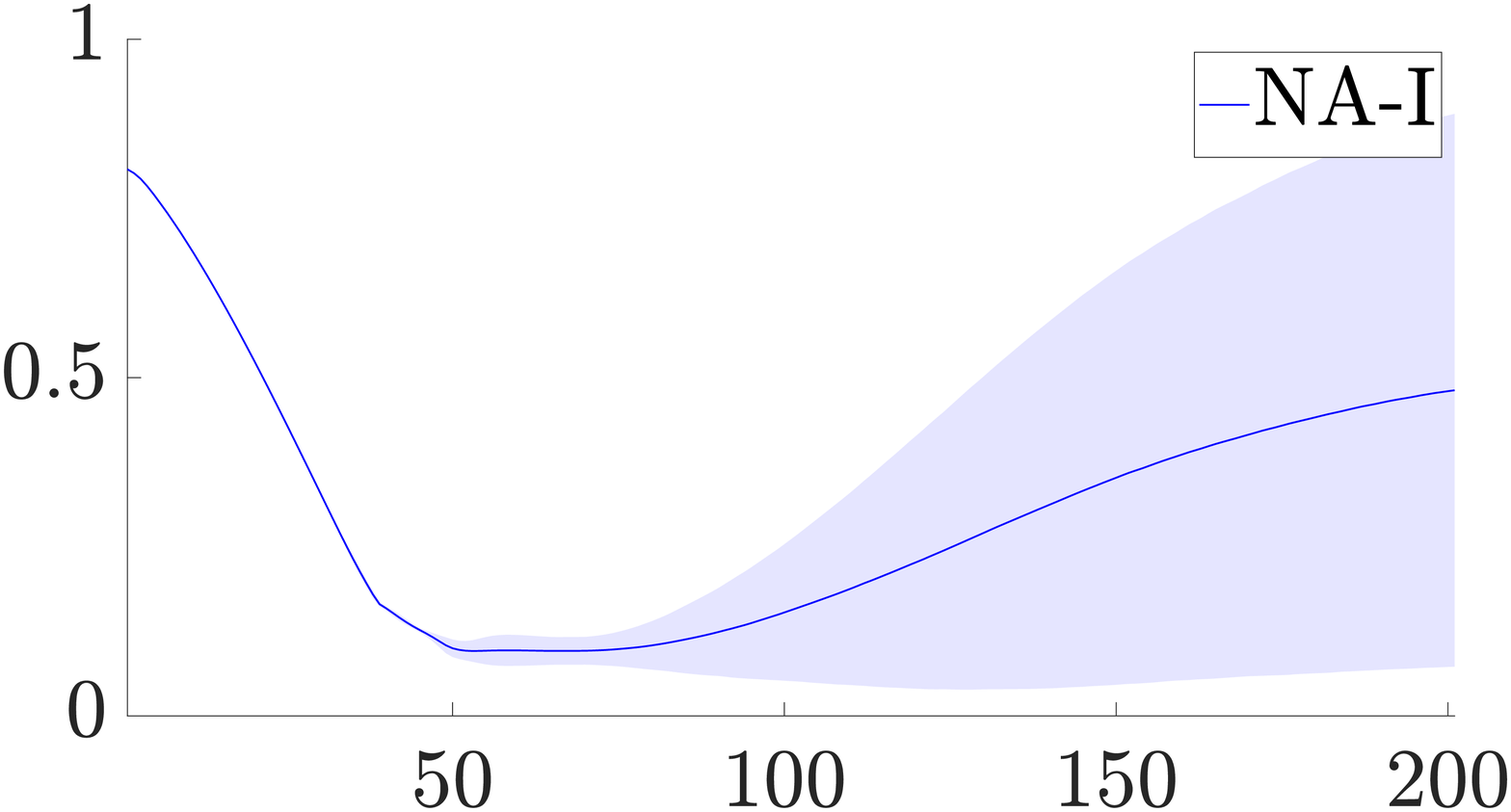}
    \end{subfigure}
    \caption{The average reaching error and its standard deviation in trials versus time step in the simulation test.}
    \label{fig:sim_test}
\end{figure*}

\subsection{The Test Study in Simulation}

The simulation test trials of the robot end-effector reaching error $E^{\mathrm{e}}_t$ of the five agents in Sec.~\ref{sec:training} are illustrated in Fig.~\ref{fig:sim_test}. The lines in the figure show the averaged reaching error $E^{\mathrm{e}}_t$ over all 500 trials and the shadow region depicts the standard deviation over all trials. It is noticed that the NA-I agent achieves the worst test performance reflected by its large average error (the dark blue line) and large variance (the shadow blue region). This indicates the bad robustness of the conventional RL agent with the existence of kinematic model deviations. Meanwhile, the NA-P agent shows a small ultimate reaching error with a very small variance. This can be recognized as a baseline representing the highest possible test performance of a conventional RL agent since it is trained and tested on an identical model. It is also noticed that both KRA-IS and KORA agents achieve similar test performance to NA-P, which indicates that both of them are sufficiently robust to the model deviations to provide an equivalent effect to the high-performance baseline NA-P. Besides, the KRA agent, with the less extent of randomization compared to KRA-IS and KORA, has a larger variance and a slightly bigger ultimate reaching error. Note that KRA-IS and KORA are subject to the same extent of randomization in the kinematic domain. Their difference is that KRA-IS has additional randomization in the time domain driven by the RT-IS, while KORA is additionally randomized in the observation domain. This verifies that the robustness of KRA-IS is due to the additional randomization exerted on the time domain. Thus, the simulation test study conducted in this subsection indicates the superior robustness of the KRA-IS agent compared to the conventional RL agent (NA-I) and the conventional domain-randomized RL agent (KRA). Its advantage over the enhanced domain-randomized RL agent KORA will be reflected by the hardware experiment study in the next subsection.

\subsection{The Test Study on Robot Hardware}

For the test study on the robot hardware, we are confined by the limited time resource and the safety requirements. Therefore, we perform 50 trials on the hardware test and define two numerical metrics, namely the ultimate reaching error $E^{\mathrm{e}}_N$ and the success rate $\rho_{\mathrm{s}}$ to evaluate the test performance. The ultimate reaching error $E^{\mathrm{e}}_N$ is the value of $E^{\mathrm{e}}_t$ at the ultimate time $t=N$, in meters. The success rate is defined as $\rho_{\mathrm{s}} = N_{s}/N_{\mathrm{ttl}}$ in percentage, where $N_{s}$ is the number of trials of which the ultimate error $E^{\mathrm{e}}_N < \varepsilon$, and $N_{\mathrm{ttl}}=50$ is the total number of the trials. The values of the numerical metrics of the 50 test trials on robot hardware are presented in Tab.~\ref{tab:impresult}.

\linespread{1.2}
\begin{table}[htbp]
\caption{Numerical Metrics of the Test on Hardware}
    \centering
    \begin{tabular}{c|c|c|c|c|c}
    \hline
        \textbf{Agent} & KRA-IS & KRA & KORA & NA-P & NA-I \\
    \hline
        $\rho_{\mathrm{s}}$ & 94 & 0 & 0 & 80 & 0 \\
    \hline
        $E^{\mathrm{e}}_N$ & 0.0367 & 0.0916 & 0.0692 & 0.0432 & 0.1530 \\
    \hline
    \end{tabular}
    \label{tab:impresult}
\end{table}
\linespread{1}

From Tab.~\ref{tab:impresult}, we notice the obvious advantage of the KRA-IS over the rest of the agents in terms of both the success rate and the ultimate reaching error. It achieves the highest success rate of 94\% and the lowest reaching error of $0.0367\,$m. Note that its performance is followed by NA-P, the conventional RL agent trained on the precise robot model, which is expected to present good test performance on the real robot, with a decent success rate of 80\% and reaching an error of $0.0432\,$m. The fact that NA-P loses performance compared to KRA-IS indicates that the latter has better generalizability than the conventional RL agents. Even though the model used to train NA-P is from Kinova\textregistered and represents the most precise identification of the robot system, sim-to-real uncertainties still exist which is not reflected by the simulation test study.  This comparison study shows that KRA-IS has superior robustness compared to the conventional domain-randomization approach.

{It is within our expectation that the NA-I trained using an imprecise robot model performs the worst among all agents with zero success rate and the largest ultimate error $E^e_N$. This reflects the lack of robustness of the conventional RL agent. It is worth mentioning that the mismatch was included in the randomization range of KRA and KORA (Sec.~\ref{sec:dra}), and both agents worked relatively better than NA-I.} Nevertheless, it is surprising that the domain-randomized agents, KRA and KORA, also achieve zero success rate, even though they have smaller ultimate errors than the NA-I. This means that the ultimate reaching errors of all their trials are larger than the predefined threshold of $0.05\,$m. We believe this is due to the improper extent of randomization which makes the agents not robust enough to the sim-to-real uncertainties. Also, it is noticed that KORA which has additional randomization in the observation domain than KRA, has a smaller ultimate error $0.0692\,$m than the latter $0.0916\,$m. This validates the superiority of KORA over KRA due to its better generalizability. 

{Therefore, we can conclude that involving randomization or stochasticity helps improve the robustness of the agent against the sim-to-real uncertainties. {Naturally, the stochasticity of the simulation hardware is case-specific and depends on the configuration of the setup. Therefore, to further guarantee the efficiency of RT-IS, using a wider range of computational hardware is necessary. However, this was unfortunately unavailable to the authors at this point. In addition, more comprehensive task definitions for manipulation scenarios are also needed to better support our conclusion.} However, to what extent the model domains should be randomized is a tricky problem and is difficult to determine. If not properly determined, the randomized agents could show inferior performance like the KRA and KORA. However, utilizing the RT-IS greatly improves the robustness of the agent. The presented comparison study indicates that it is even more robust than the conventional RL agent trained using the precise robot model. Since the real-time simulation mode mainly randomizes the time domain, we can also conclude that the time domain randomization is more effective than the domains of kinematics and observation, a feature also present in methods such as in~\cite{bouteiller2021reinforcement}.}

\section{Conclusion}\label{sec:con}
{In this paper, we investigate the stochastical property of the intrinsic stochasticity of real-time simulation (RT-IS) and explore the feasibility of utilizing it to facilitate the transfer of an RL robot controller from simulation to the real world. Significance tests in multiple robot tasks have verified its correlation with the hardware computation components. The variability range and the RMS stochasticity value are also assessed and quantified as opposed to that of a real-world robot. At this stage, it was concluded that the computational hardware utility is correlated to the real-time simulation stochasticity. Then, the stochastic real-time simulation (RT-IS) was used in an RL framework for an essential point-to-point (P2P) task. {Three basic domain randomized agents are designed for this experiment, one of which is powered by RT-IS. Compared to the other agents, the RT-IS-powered agent achieves the highest success rate and lowest reaching errors. This shows that by enabling RT-IS in RL training, a straightforward domain randomized model can readily be transferred to real environments. Future work in this regard should be conducted to verify this phenomenon for more advanced methods and baselines.} At this stage, we can deduce that stochastic simulation improves generalizability and eases the sim-to-real. We can summarize the conclusions in these two points:
\begin{enumerate}
    \item Considering both stages of the experiments, taking advantage of the simulation's computational hardware is beneficial to the robustness of RL agents. This is a physically plausible source of noise in simulation engines that emulates the physical setup's noise. {The correlation between computational hardware and RT-IS suggests a plausible origin for the stochasticity in the simulation environment. It also  shows a promise for future work to manipulate the hardware resources such that we can exploit RT-IS at a favorable stochasticity level.}
    \item Even independent from the role of computational hardware, RT-IS can help train RL agents to be less sensitive to sim-to-real, with fewer heuristics, no task-dependency, and better generalizability. Activating this feature is straightforward and does not require additional design or manipulation by an expert. To the best of the authors' knowledge, this level of automation for such improvement stands out from the related literature.
\end{enumerate}
The price we pay for such generalizability is the decreased training performance, which, however, is a worthwhile trade-off in most practical applications. Also, while our proposed method successfully produces a robust RL agent, it is important to note that using real-time simulations for training RL methods can be more time-consuming than using non-real-time simulations. This is because non-real-time simulations allow for the specification of time steps that can significantly speed up a simulation by huge magnitudes. More efficient methods to train RL agents powered by RT-IS will be investigated in future work. We will also continue to explore the application of these agents to more complicated robot manipulation tasks.}

\bibliographystyle{IEEEtran}
\bibliography{IEEEabrv, reference.bib}

\end{document}